\documentclass[letterpaper]{article} % DO NOT CHANGE THIS
\usepackage{aaai24}  % DO NOT CHANGE THIS
\usepackage{times}  % DO NOT CHANGE THIS
\usepackage{helvet}  % DO NOT CHANGE THIS
\usepackage{courier}  % DO NOT CHANGE THIS
\usepackage[hyphens]{url}  % DO NOT CHANGE THIS
\usepackage{graphicx} % DO NOT CHANGE THIS
\usepackage{natbib}  % DO NOT CHANGE THIS AND DO NOT ADD ANY OPTIONS TO IT
\usepackage{caption} % DO NOT CHANGE THIS AND DO NOT ADD ANY OPTIONS TO IT
\usepackage{algorithm}
\usepackage{algorithmic}
\usepackage{multicol}
\usepackage{newfloat}
\usepackage{listings}
\DeclareCaptionStyle{ruled}{labelfont=normalfont,labelsep=colon,strut=off} % DO NOT CHANGE THIS
\floatstyle{ruled}
\newfloat{listing}{tb}{lst}{}
\floatname{listing}{Listing}
\usepackage{algorithm}
\usepackage{algorithmic}
\usepackage{booktabs}
\newcommand{\tref}{Table~\ref}
\newcommand{\fref}{Fig.~\ref}
\usepackage{multirow}
\usepackage{amssymb}
\usepackage{amsmath, bm}
\usepackage{subcaption}
\usepackage[table,xcdraw]{xcolor}

\title{Agile Multi-Source-Free Domain Adaptation}
\author{
    Xinyao Li\textsuperscript{\rm 1},
    Jingjing Li\textsuperscript{\rm 1,2}\thanks{Corresponding author},
    Fengling Li\textsuperscript{\rm 3},
    Lei Zhu\textsuperscript{\rm 4},
    Ke Lu\textsuperscript{\rm 1}
}
\affiliations{
    \textsuperscript{\rm 1}University of Electronic Science and Technology of China (UESTC)\\
    \textsuperscript{\rm 2}Shenzhen Institute for Advanced Study, UESTC \\
    \textsuperscript{\rm 3}University of Technology Sydney\\
    \textsuperscript{\rm 4}School of Electronic and Information Engineering, Tongji University \\
    xinyao326@outlook.com, lijin117@yeah.net, \{fenglingli2023, leizhu0608\}@gmail.com, kel@uestc.edu.cn
}

\usepackage{bibentry}
\begin{document}

\maketitle

\begin{abstract}
Efficiently utilizing rich knowledge in pretrained models has become a critical topic in the era of large models. This work focuses on adaptively utilizing knowledge from multiple source-pretrained models to an unlabeled target domain without accessing the source data. Despite being a practically useful setting, existing methods require extensive parameter tuning over each source model, which is computationally expensive when facing abundant source domains or larger source models. To address this challenge, we propose a novel approach which is free of the parameter tuning over source backbones. Our technical contribution lies in the Bi-level ATtention ENsemble (Bi-ATEN) module, which learns both intra-domain weights and inter-domain ensemble weights to achieve a fine balance between instance specificity and domain consistency. By slightly tuning source bottlenecks, we achieve comparable or even superior performance on a challenging benchmark DomainNet \textbf{with less than 3\% trained parameters and 8 times of throughput compared with SOTA method}. Furthermore, with minor modifications, the proposed module can be easily equipped to existing  methods and gain more than 4\% performance boost. Code is available at  https://github.com/TL-UESTC/Bi-ATEN.
\end{abstract}

\section{Introduction}

Large-scale models have drawn significant attention for their remarkable performance across a spectrum of applications~\cite{ramesh2022hierarchical,irwin2022chemformer,lee2020biobert}. Considering that training large models from scratch requires tremendous computational costs, fine-tuning has become a predominant approach to  transfer knowledge from large pretrained models to downstream tasks~\cite{long2015learning,guo2020adafilter}. However, this paradigm heavily relies on labeled training data and suffers from significant performance decay when target data exhibits distribution shift from pretraining data~\cite{ben2010theory}. Moreover, we usually have multiple pretrained models trained on different sources or architectures on hand, e.g., medical diagnostic models trained on distinct regions or patient groups.  Demands to maximally utilizing knowledge from multiple pretrained models are common in real world applications. To this end, Multi-Source-Free Domain Adaptation~(MSFDA)~\cite{ahmed2021unsupervised,dong2021confident} emerges as a promising technique to address  these challenges by enabling holistic adaptation of multiple pretrained source models to an unlabeled target domain, while not accessing source training data.

\begin{table}[t]
    
    \centering
    
    \small
    \begin{tabular}{@{}ccccc@{}}
    \toprule
    Method & Param. & Backbone & Acc. & Throughput  \\ \midrule
    CAiDA                   & 120.2M                                                                         & ResNet50                  & 46.8                  & 91                                                                                    \\
    PMTrans                 & 447.4M                                                                         & Swin                      & 59.1                  & 46                                                                                    \\
    ATEN (ours)             & 4.9M                                                                           & Swin                      & 59.1                  & 970                                                                                   \\
    Bi-ATEN (ours)         & 10.6M                                                                          & Swin                      & 59.6                  & 369                                                                                   \\ \bottomrule
\end{tabular}
\caption{Computation overhead and performance comparison between different methods on DomainNet.}
\label{tab1}
\end{table}

Existing  MSFDA methods~\cite{ahmed2021unsupervised,dong2021confident,han2023discriminability,shen2023balancing} typically tackle the problem via a two-step framework, i.e., \textit{(1)} Tune each source model thoroughly towards target domain, and \textit{(2)} Learn source importance weights  to assemble the source models. However, their overwhelming limitations in  computational efficiency and scalability prevent their  applications on large-scale problems. For \textit{step (1)}, the number of models to tune increases linearly along with the number of source domains, which could become unacceptable for large-scale problems with abundant source domains. The necessity of tuning all parameters for each model also makes it infeasible to scale up these methods to larger models.
In \tref{tab1} we compare the performance and trainable parameters of CAiDA~\cite{dong2021confident}, PMTrans\footnote[1]{PMTrans is a single-source domain adaptation method and we evaluate it on MSFDA setting by taking its single-best results.}~\cite{zhu2023patch} and our methods on a challenging benchmark DomainNet~\cite{peng2019moment} with 6 domains. As a typical MSFDA framework, CAiDA performs poorly due to limited performance of ResNet-50~\cite{he2016deep} backbone. By equipping a stronger backbone SwinTransformer~\cite{liu2021swin}, a potential performance boost of +12.3\% is achieved at a cost of four times of parameters to tune.
On the other hand, we aim to achieve superior performance by equipping SwinTransformer while demanding significantly less training cost, presenting a more feasible and agile solution for MSFDA on large models.
For \textit{step (2)}, current MSFDA methods learn \textbf{domain-level} ensemble weights, applying identical ensemble strategy across all target instances. Although the learned weights are intuitively interpretable in terms of domain transferablity, they unavoidably introduce misalignment and bias at instance-level. This controversy inherently introduces a trade-off between instance specificity and domain consistency of ensemble weights, which has not been well exploited by existing methods.

\begin{figure}[t]
    \centering
    \begin{subfigure}{0.22\textwidth}
        \includegraphics[width=\textwidth]{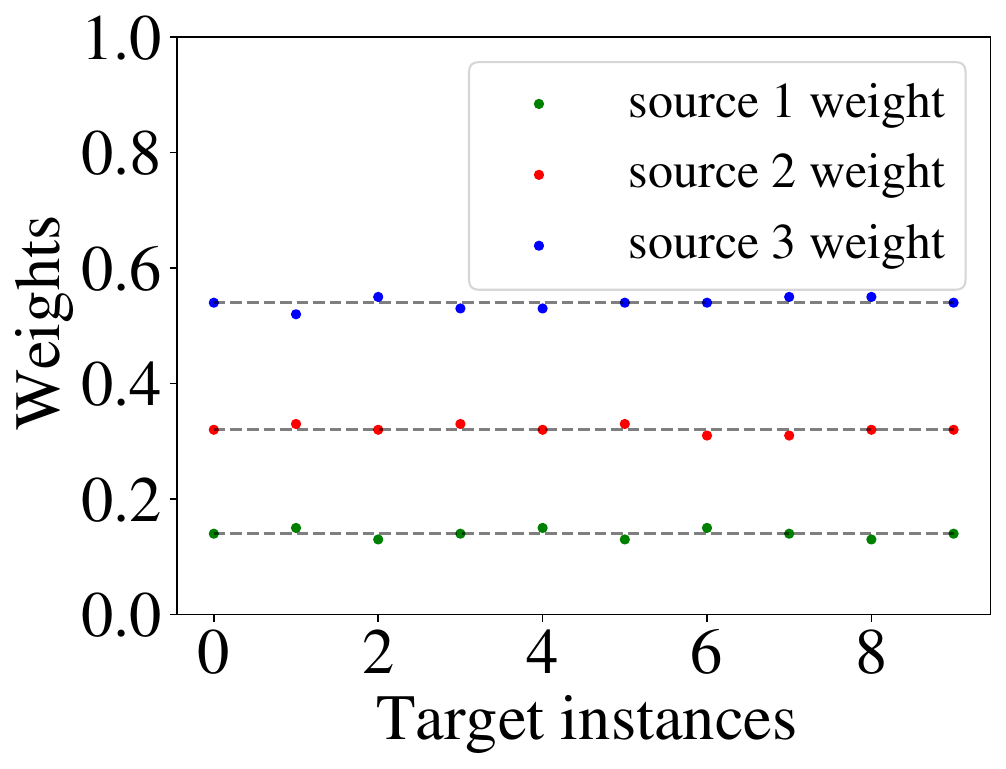}
        \caption{without instance specificity}
        \label{fig1_1}
    \end{subfigure}
    \hfill
    \begin{subfigure}{0.22\textwidth}
        \includegraphics[width=\textwidth]{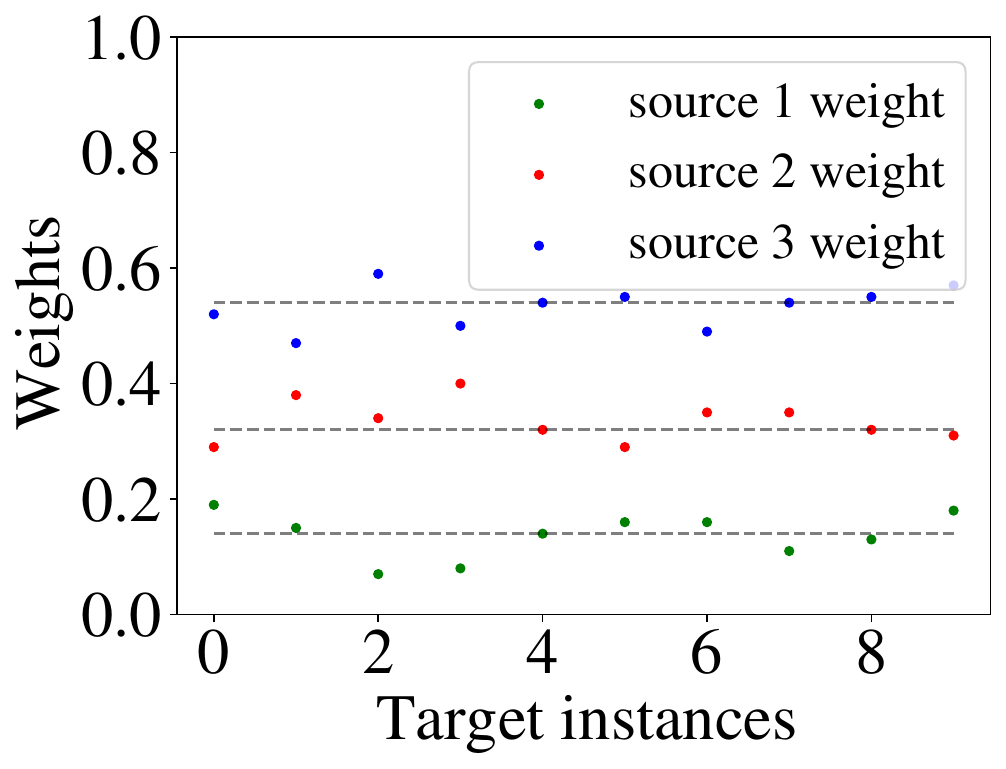}
        \caption{with instance specificity}
        \label{fig1_2}
    \end{subfigure}
    \caption{Illustration of instance specificity and domain consistency. Dots are weights assigned to each target sample.}
    \label{fig1}
\end{figure}

Recent success of model ensemble methods~\cite{shu2021zoo,shu2022hub} suggests that it is effective to transfer knowledge by designing adaptive ensemble weights. While optimal strategies are hard to learn~\cite{mohammed2023comprehensive}, we resort to slight tuning of several domain-specific bottleneck layers, costing less than 0.1\% of tuning the whole model. 
As stated above, the key to designing effective weights is to exploit both domain-level transferabilities and instance-level individual characteristics, as illustrated by \fref{fig1}. Existing MSFDA methods learn weights solely from feature representations, neglecting the potential transferability mismatch between features and outputs, i.e., transferable target features do not always lead to accurate predictions. To address this issue, we propose to  introduce additional semantic information from classifiers for deriving weights. For each feature representation, we first learn \textit{intra-domain weights} to mitigate transferability mismatch by finding the most compatible classifier that produces unbiased outputs. With unbiased outputs from the selected source classifier, we further learn \textit{inter-domain ensemble weights} that combine  source outputs into the final result.  We propose a novel Bi-level ATtention ENsemble~(Bi-ATEN) to effectively learn the two  weights through attention mechanisms. Bi-ATEN is capable of tailoring its ensemble decisions to the particularities of each instance, while maintaining the broader transferability trends that are consistent across domains. This balance is essential for accurate domain adaptation, where a model needs to leverage domain-specific knowledge without losing the overarching patterns that drive adaptation.

The proposed Bi-ATEN can be simplified into inter-domain ATtention ENsemble (ATEN) and plugged into existing MSFDA methods by replacing their weight-learning module. Although leaning towards domain consistency in the specificity-consistency balance, ATEN still exhibits clear performance boost over baseline methods, proving the efficacy of our design.
In a nutshell, we achieve adaptation primarily by assuring instance specificity and domain consistency  along with slight tuning of bottlenecks. \tref{tab1} provides comprehensive comparison between our methods and existing methods. Our contributions can be summarized as: (1) We propose a novel framework to agilely handle MSFDA  by learning fine-grained domain adaptive ensemble strategies. (2) We design an effective module Bi-ATEN that learns both intra-domain weights and inter-domain ensemble weights. Its light version ATEN can be equipped to existing MSFDA methods to boost performance. (3) Our method significantly reduces computational costs while achieving state-of-the-art performance, making it feasible for real-life transfer applications  with large  source-trained models. (4) Extensive experiments on three challenging benchmarks and detailed analysis demonstrates the success of our design.

\section{Related Work}
\textbf{Source-free domain adaptation} (SFDA) assumes no labeled source data but a source-trained model is available for  adaptation~\cite{li2021divergence}. SHOT \cite{liang2020we} pioneers the problem by proposing a clustering algorithm for pseudo-labeling and utilizes information maximization loss. Several works~\cite{ma,yang2021exploiting} follow the research line to improve or develop new clustering methods. \citeauthor{kundu2022balancing} \shortcite{kundu2022balancing} reveal insight on discriminability and transferability trade-offs and propose to mix-up original and corresponding translated generic samples to improve performance. Other relevant settings including source-free active domain adaptation~\cite{li2022source} and imbalanced SFDA~\cite{li2021imbalanced} have also been explored. 

\textbf{Multi-source domain adaptation} (MSDA) assumes that labeled source data from multiple domains are available, and tries to transfer simultaneously towards target domain with theoretical guarantees from pioneering works~\cite{ben2010theory,crammer2008learning}. M$^{3}$SDA~\cite{peng2019moment} provides theoretical insights that all source-target and source-source pairs should be aligned to achieve adaptation. DRT~\cite{li2021dynamic} proposes a dynamic module that adapts model parameters according to samples. ABMSDA~\cite{zuo2021attention} proposes a Weighted Moment Distance to ensure higher attention among more related domains. STEM~\cite{nguyen2021stem} generates a teacher-student framework to close the gap between source and target distributions. 

\begin{figure*}[t]
    \centering
    \includegraphics[width=2\columnwidth]{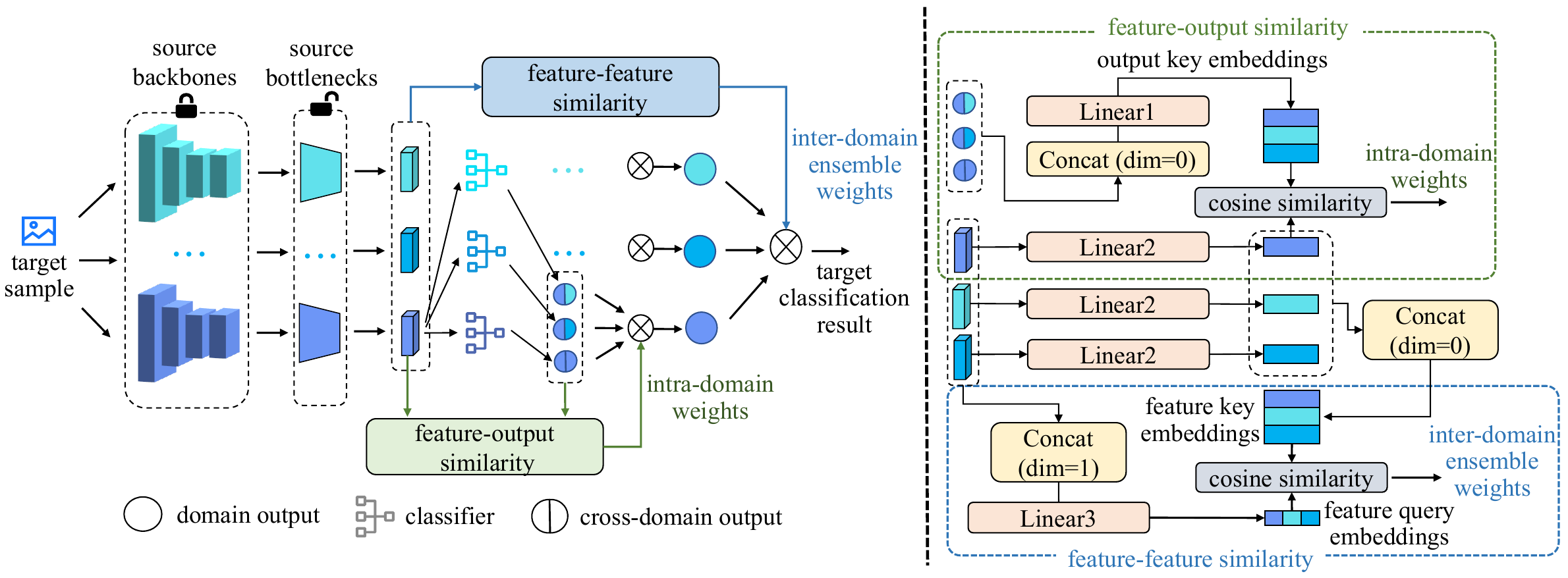} 
    \caption{Framework of our method. Different colors represent different source domains. For cross-domain outputs, colors on the left semicircles represent domains of bottleneck features while that on the right semicircles represent domains of classifiers that generate the cross-domain output. Best viewed in color.}
    \label{fig2}
\end{figure*}

\textbf{Multi-source-free domain adaptation} (MSFDA) combines SFDA and MSDA, aiming to learn optimal source model combinations that perform best on unlabeled target data. DECISION~\cite{ahmed2021unsupervised} first explores the problem and proposes to assemble source outputs with learnable weights while updating source models  via weighted information maximization. CAiDA~\cite{dong2021confident} proposes to use a similar framework but with a confident-anchor-induced pseudo label generator. \citeauthor{shen2023balancing} \shortcite{shen2023balancing} develop a generalization bound on MSFDA that reveals an inherent bias-variance trade-off. A  hierarchical framework is further proposed to balance the trade-off. DATE~\cite{han2023discriminability} evaluates source transferabilities via a Bayesian perspective before quantifying the similarity degree by a multi-layer perception. All forementioned methods learn domain-level importance  regardless of instance characteristics, which unavoidably limits their performance.

\section{Method}
\subsection{Problem Definition}
Assume we have $n$ source-trained models $\{h_s^i\}_{i=1}^{n}$ for $C$-category classification task.
Given an unlabeled target domain $\{X_t\}$ with identical categories, the goal is to optimize all $n$ source models towards satisfactory performance on the  target domain. Following~\cite{tzeng2014deep}, a bottleneck layer $k_s$ with parameter $\theta _{k_s}$ is applied after the feature extractor $f_s$ with parameter $\theta _{f_s}$, and before the final fully-connected classifier $g_s$ with parameter $\theta _{g_s}$. Given a target sample $x_t$, we define its bottleneck feature with $d_k$ dimensions produced by source model $h_s^i$  as $\phi_t^i = (k_s^i  \circ  f_s^i)(x_t)$, and the output of source model $h_s^i$ can be denoted as $y_t^i=g_s^i(\phi_t^i)$. Specifically, in this paper we consider cross-domain outputs obtained by forwarding $\phi_t^i$ through a classifier from another domain $j$, i.e., $y_t^{ij}=g_s^j(\phi_t^i)$.  By learning intra-domain weights  $ \boldsymbol{\alpha^i} $, unbiased domain output for feature $\phi_t^i$ is denoted as $\tilde{y}_t^i=\sum _{j=1}^{n} \alpha_j^i y_t^{ij}$. Inter-domain ensemble weights $\boldsymbol{\beta} $ are further learned to obtain final output $\ddot{y} _t=\sum _{i=1}^{n} \beta_i \tilde{y}_t^i$. Our goal is to learn optimal $ \{\boldsymbol{\alpha^i}\}_{i=1}^n $, $\boldsymbol{\beta} $ and bottleneck parameters $\theta _{k_s}$  that minimizes training loss.

\subsection{Overview}
\fref{fig2} depicts our  framework. A target sample is forwarded through the   source models to extract the bottleneck features. Instead of directly generating outputs by specific source classifier, we compute all possible cross-domain outputs with respect to current feature by forwarding it through all source classifiers. Intra-domain weights $ \{\boldsymbol{\alpha^i}\}_{i=1}^n $ are computed between the feature representation and all output vectors for obtaining unbiased outputs. Subsequently, inter-domain  weights $\boldsymbol{\beta}$ are learned to assemble the unbiased domain  outputs into the final classification result. Note that both source backbones and source classifiers remain frozen during the entire training process. Laying at the core of the framework is the  Bi-ATEN module, as depicted on the right of \fref{fig2}. It simultaneously learns $ \{\boldsymbol{\alpha^i}\}_{i=1}^n $ from feature-output similarities and $\boldsymbol{\beta}$ from feature-feature similarities. Next we elaborate on the detailed design of each module.

\subsection{Bi-level Attention Ensemble}
\textbf{Intra-domain weights.} 
All current MSFDA methods adopt an end-to-end training paradigm that treats each source model as a whole~\cite{dong2021confident}. However, the distribution shifts between target and source data can lead to mismatches within the source model components like bottlenecks and classifiers. Inspired by deep model reassembly methods~\cite{yang2022deep}, we propose to improve current MSFDA paradigms by performing a partial model reassembly.
We explore compatible  bottleneck-classifier pairs tailored towards target data characteristics, and obtain the reassembled result  by summing over weighted cross-domain outputs of bottleneck-classifier pairs.
Given bottleneck feature from the $i_{th}$ source domain $\phi_t^i \in \mathbb{R}^{d_k}$, we first obtain its cross-domain outputs by:
\begin{equation}
    \label{eq1}
    O_t^i = Concat(\{ \theta_{g_s}^j \phi_t^i \}_{j=1}^n, \ \dim=0),
\end{equation}
where $O_t^i \in \mathbb{R}^{n \times C}$ is  cross-domain output matrix for the $i_{th}$ feature. Since source classifier parameters are fixed, our aim to find the most compatible classifier can be converted to finding the most similar output vector after classification linear transformation $\theta_{g_s}$. We adopt cosine similarity to eliminate norm mismatch between features and outputs:
\begin{equation}
    \label{eq2}
    Sim_t^i = Cosine(\phi_t^i W^{F}, O_t^i W^{O}),
\end{equation} 
where $Sim_t^i \in \mathbb{R}^{n}$ is similarity vector, $W^{F} \in \mathbb{R}^{d_k\times d_{emb}}$ (Linear2 in \fref{fig2}) and $W^{O}\in \mathbb{R}^{C\times d_{emb}}$ (Linear1 in \fref{fig2}) are linear transforms that transform feature and output into the same embedding dimension $d_{emb}$. Then, intra-domain weights are obtained by applying softmax operation over the similarity vector:
\begin{equation}
    \label{eq3}
    \boldsymbol{\alpha^i} = Softmax(Sim_t^i).
\end{equation}
Finally, assembled output for domain $i$ is obtained by:
\begin{equation}
    \label{eq4}
    \tilde{y}_t^i=\sum _{j=1}^{n} \alpha_j^i \theta_{g_s}^j \phi_t^i.
\end{equation}

We regard output $\tilde{y}_t^i$ as unbiased if it is: (1) Confident. Ambiguous outputs imply multiple possible interpretations on the feature, increasing the risk of feature-output mismatch. (2) Diverse. Overly consistent classification results lead to mode collapse where certain classes are rarely considered. We apply IM loss~\cite{liang2020we}, a base component shared by current MSFDA methods, to assure unbiased intra-domain ensemble:
\begin{equation}
    \label{eq5}
    \mathcal{L}_{intra} = \sum_{i=1}^{n} \mathcal{L}_{IM}(Softmax(\tilde{y}_t^i)),
\end{equation}
where $\mathcal{L}_{IM}$ is defined as:
\begin{align}
    \label{eq6}
    & \mathcal{L}_{IM}(y)=\mathcal{L}_{ent}(y)-\mathcal{L}_{div}(y), \rm{where} \\
    & \mathcal{L}_{ent}(y)=-\mathbb{E}_{x_t \in X_t}\left[\sum_{c=1}^C \delta_c(y) \log  \delta_c(y)\right], \notag \\
    & \mathcal{L}_{div}(y)=-\sum_{c=1}^{C}\bar{p}_c \ log \ \bar{p}_c, \notag
\end{align} 
where $\bar{p}_c=-\mathbb{E}_{x_t \in X_t}\delta_c(y)$ and $\delta_c(\cdot )$ takes the $c_{th}$ logit.
\textbf{Inter-domain weights.} We derive ensemble weights from bottleneck features. Motivated by the success of attention mechanism~\cite{vaswani2017attention}, we obtain inter-domain weights by computing attention between different linear representations of bottleneck features.  To allow intra-domain adjustments according to inter-domain weights, the  transform matrix $W^F$ is shared with that in Eq. (2):
\begin{equation}
    \label{eq7}
    \hat{\phi}_t^K = Concat(\{\phi_t^i W^F\}_{i=1}^n, \ \dim=0).
\end{equation}
For query embeddings, features are first concatenated before linearly transformed:
\begin{equation}
    \label{eq7}
    \hat{\phi}_t^Q = Concat(\{\phi_t^i\}_{i=1}^n, \ \dim=1)W^{QF},
\end{equation}
where $W^{QF}\in \mathbb{R}^{(nd_k) \times d_{emb}}$ is the query transform matrix (Linear3 in \fref{fig2}). Similar to intra-domain weights, we compute inter-domain weights via:
\begin{equation}
    \label{eq8}
    \boldsymbol{\beta} = Softmax(Cosine(\hat{\phi}_t^Q, \hat{\phi}_t^K)).
\end{equation}
Final ensemble result is then obtained by:
\begin{equation}
    \label{eq9}
    \ddot{y}_t=\sum _{i=1}^{n} \beta_i \tilde{y}_t^i.
\end{equation}
Apart from being confident and diverse, the final ensemble result should more importantly be correct. Since no label is available, in this work we adopt a \textbf{dynamic-cluster-based strategy} to provide pseudo labels for classification. The dynamic is two-fold: dynamic feature combinations and dynamic centroids for each instance. We first compute centroid for class $c$ generated by source model $h_s^i$ by:
\begin{equation}
    \label{eq10}
    \mu_c^i = \frac{\sum_{x_t\in X_t} \delta_c(Softmax(\ddot{y}_t))\phi_t^i}{\sum_{x_t\in X_t} \delta_c(Softmax(\ddot{y}_t))},  
\end{equation}
where $\delta_c(\cdot )$ takes the $c_{th}$ logit. Dynamic centroid for the $m_{th}$ target sample $x_t^m$ of class $c$ is computed by assembling all centroids using instance-specific inter-domain weight $\boldsymbol{\beta^m}$:
\begin{equation}
    \label{eq11}
    \tilde{\mu}_c^m = \sum_{i=1}^n \beta_i^m \mu_c^i.
\end{equation}
For target samples, their feature representations are dynamically obtained by assembling all source bottleneck features:
\begin{equation}
    \label{eq12}
    \tilde{\phi}_t^m=\sum_{i=1}^{n} \beta_i^m \phi_t^{mi} , 
\end{equation}
where $\phi_t^{mi}$ is bottleneck feature extracted by source model from domain $i$ for sample $x_t^m$.
Finally, we generate pseudo label for $x_t^m$ by:
\begin{equation}
    \label{eq13}
    y_t = \arg \underset{c}{\max} \; Cosine(\tilde{\phi}_t^m, \ \tilde{\mu}_c^m).
\end{equation}
Dynamic clustering greatly extends the diversity and flexibility of generated pseudo labels. As Bi-ATEN becomes more reliable, quality of pseudo labels is concurrently improved, which in turn helps the training of Bi-ATEN. With pseudo labels, objective for final  output is formulated as:
\begin{equation}
    \label{eq14}
    \mathcal{L}_{inter} = \gamma CE(\ddot{y}_t, y_t) + \mathcal{L}_{IM}(Softmax(\ddot{y}_t)),
\end{equation}
where $\gamma$ is a hyperparameter and $CE(\cdot)$ is cross entropy loss with label smoothing~\cite{szegedy2016rethinking}. Overall objective is given as:
\begin{equation}
    \label{eq15}
    \mathcal{L} = \mathcal{L}_{inter} + \lambda \mathcal{L}_{intra},
\end{equation}
where $\lambda$ is a trade-off hyperparameter. We train our model by solving the following optimization problem:
\begin{equation}
    \label{eq16}
    \boldsymbol{\alpha }, \ \boldsymbol{\beta }, \ \theta_{k_s} = \arg\min \;\mathcal{L}.
\end{equation}

\begin{table*}[t]
    \centering
    \small
    
    %\resizebox{0.9\linewidth}{!}{
    \begin{tabular}{c|c|c|ccccccccc}
    \toprule
    Method     & SF        & Backbone     & $\rightarrow$clp           & $\rightarrow$inf           & $\rightarrow$pnt           & $\rightarrow$qdr           & $\rightarrow$rel           & $\rightarrow$skt           & Avg.    & Param. & Train time     \\ \midrule
    M$^3$SDA & $\times $ & \multirow{4}{*}{ResNet101}  & 58.6          & 26.0          & 52.3          & 6.3          & 62.7          & 49.5          & 42.6   &    42.48M  & / \\
    LtC-MSDA & $\times $ &   & 63.1          & 28.7          & 56.1        & 16.3          & 66.1          & 53.8          & 47.4    & 42.50M  & /   \\
    STEM         & $\times $ &  & 72.0          & 28.2          & 61.5          & 25.7          & 72.6          & 60.2          & 53.4    &43.78M  &/    \\
    DRT          & $\times $ &  & 71.0          & 31.6       & 61.0          & 12.3          & 71.4          & 60.7          & 51.3   &60.90M     &/  \\ \midrule
    DECISION     & $\surd $  &  \multirow{5}{*}{ResNet50}    & 61.5          & 21.6          & 54.6       & 18.9       & 67.5       & 51.0       & 45.9    &120.14M    & 2.9H  \\
    DATE & $\surd$ &  & 61.2 & 22.7 & 53.5 & 18.1 & 69.8 & 50.9 & 46.0 & / & / \\
    CAiDA        & $\surd $  &     & 63.6          & 20.7          & 54.3          & 19.3          & 71.2          & 51.6          & 46.8 &120.20M    & 3.0H     \\
    Surrogate    & $\surd $  &     & 66.5          & 21.6          & 56.7          & 20.4          & 70.5          & 54.4          & 48.4  & /   & /     \\
    TransMDA    & $\surd $ &              & 71.7        & 29.0         & 61.4       & 18.6         & 74.1     & 60.9    &  52.6   & /   & /  \\\midrule
    CDTrans-best & $\times $ & DeiT-base    & 69.0        & 31.0        & 61.5        & 27.2         & 72.6          & 58.1          & 53.2  &428.23M    & /   \\ \midrule
    
    SSRT-best    & $\times $ &   ViT-base    & 70.6          & 37.1        & 66.0          & 21.7          & 75.8          & 59.8          & 55.2   &442.74M    & /   \\ \midrule
    DRT     & $\times $  & \multirow{5}{*}{SwinTransformer} & 74.6          & 33.2          & 64.8        & 20.3        & 76.4       & 64.6     & 55.6      &91.43M & /    \\
    AVG-ENS          & $\surd $  &       & 74.1          & 35.3          & 66.1          & 15.0          & 81.6          & 62.9          & 55.8 & /      & /   \\
    PMTrans-best & $\times $ &           & 74.1          & 35.3          & \textbf{70.7} & \textbf{30.9} & 79.8          & 63.7          & 59.1  &447.43M   & /     \\
    \textbf{ATEN (ours)}       & $\surd $  &      & 76.6          & 37.2          & 68.6          & 24.0          & 83.5          & 64.6          & 59.1   &\textbf{4.92M}   &\textbf{0.6H}    \\
    \textbf{Bi-ATEN (ours) }   & $\surd $  &      & \textbf{77.0} & \textbf{38.5} & 68.6        & 25.0        & \textbf{83.6} & \textbf{64.9} & \textbf{59.6}&10.56M & 1.2H \\ \bottomrule
    \end{tabular}
    \caption{Results on DomainNet. SF denotes whether the method follows source-free setting. Best results are  in bold font.}
    \label{tab2}
\end{table*}

\subsection{Attention Ensemble as a Pluggable Module} 
Consider an extreme situation where $\boldsymbol{\alpha^i}$ contains a single one at the $i_{th}$ location and zeros elsewhere. It simplifies Bi-ATEN to ATEN with only inter-domain ensemble weights $\boldsymbol{\beta}$, which aligns with weight learning paradigm of existing MSFDA methods, and can therefore replace their weight learning module easily. Assume objective of the original MSFDA method as $\mathcal{L}_{origin}$, the optimization goal after equipping ATEN becomes:
\begin{equation}
    \label{eq18}
    \boldsymbol{\beta }, \ \theta_{k_s}, \ \theta_{f_s} = \arg\min \;\mathcal{L}_{origin}.
\end{equation}
$\boldsymbol{\alpha^i}$s are fixed as one-hot vectors as described above, thus saving the training of $W^O$.

\subsection{Training Process}
We design an alternate training procedure for Bi-ATEN. We observe that for target domains with relatively smaller domain gap, the domain-specific source classifiers already show satisfactory performance, while for those with larger domain gap, intra-domain weights are vital for adaptive feature-classifier matching. Considering both cases, in certain epochs we manually set  $\boldsymbol{\alpha^i}$ to one-hot vectors as in ATEN.  Different from Eq. (\ref{eq18}), we still update $W^O$ via Eq. (\ref{eq5}). Such alternate training utilizes the benefits of both strategies, striking a balance between intra-domain compatibility and domain-consistent adaptation.

\section{Experiments}
In this section we present main results and further analysis. Implementations are based on MindSpore and PyTorch.

\subsection{Datasets and Baselines}
\textbf{Datasets.} We evaluate our method on three MSFDA benchmarks Office-Home~\cite{officehome}, Office-Caltech~\cite{officecal} and DomainNet~\cite{peng2019moment}.
Office-Home is divided into 65 categories with 4 domains Art, Clipart, Product and RealWorld. Office-Caltech is extended from Office31~\cite{office31} by adding Caltech~\cite{caltech256} as a fourth domain. DomainNet 
%is the most challenging domain adaptation benchmark so far, 
is composed of 0.6 million samples from six distinct domains, each containing 345 categories. 

\textbf{Baselines.} On Office-Home and Office-Caltech we validate boost obtained by equipping ATEN to existing MSFDA methods: DECISION~\cite{ahmed2021unsupervised}, CAiDA~\cite{dong2021confident}, DATE~\cite{han2023discriminability}, and compare with other MSDA methods including M$^3$SDA~\cite{peng2019moment}, LtC-MSDA~\cite{lctmsda}, MA~\cite{ma}, NRC~\cite{yang2021exploiting} and SHOT~\cite{liang2020we}. Baseline results of forementioned methods  are cited from DATE. On DomainNet we compare our ATEN and Bi-ATEN against various competing baselines implemented on various backbones. ResNet101~\cite{he2016deep}: M$^3$SDA, LtC-MSDA, STEM~\cite{nguyen2021stem} and DRT~\cite{li2021dynamic}. ResNet50~\cite{he2016deep}: DECISION, CAiDA, DATE, Surrogate~\cite{shen2023balancing} and TransMDA~\cite{li2023transformer}. DeiT~\cite{deit}: CDTrans~\cite{xu2021cdtrans}. ViT~\cite{vit}: SSRT~\cite{ssrt}. SwinTransformer~\cite{liu2021swin}: PMTrans~\cite{zhu2023patch} and DRT  implemented by ourselves.

\begin{table*}[t]
    \centering
    \small
    
    %\resizebox{0.9\linewidth}{!}{
    \begin{tabular}{c|c|ccccc|ccccc}
        \toprule
        \multirow{2}{*}{Method} & \multirow{2}{*}{SF} & \multicolumn{5}{c|}{Office-home} & \multicolumn{5}{c}{Office-Caltech} \\ \cmidrule(l){3-12}    
         &  & $\rightarrow$Art & $\rightarrow$Clp & $\rightarrow$Prod & $\rightarrow$Real & Avg. & $\rightarrow$amazon & $\rightarrow$caltech & $\rightarrow$dslr & $\rightarrow$webcam & Avg. \\ \midrule
        M$^3$SDA & $\times$ & 67.2 & 63.5 & 79.1 & 79.4 & 72.3 & 94.5 & 92.2 & 99.2 & 99.5 & 96.4 \\
        LtC-MSDA & $\times$ & 67.4 & \textbf{64.1} & 79.2 & 80.1 & 72.7 & 93.7 & 95.1 & 99.7 & 99.4 & 97.0 \\
        MA & $\surd$ & 72.5 & 57.4 & 81.7 & 82.3 & 73.5 & 95.7 & 95.6 & 97.2 & \textbf{99.8} & 97.1 \\
        NRC & $\surd$ & 72.7 & 58.1 & 82.3 & 82.1 & 73.8 & \textbf{95.9} & 94.9 & 97.5 & 99.3 & 96.9 \\
        SHOT & $\surd$ & 72.2 & 59.3 & 82.9 & 82.8 & 74.3 & 95.7 & 95.8 & 96.8 & 99.6 & 97.0 \\\midrule
        %Surrogate & $\surd$ & 75.6 & 62.8 & 84.8 & \textbf{85.3} & \textbf{77.1} & / & / & / & / & / \\ \midrule
        DECISION & $\surd$ & 73.3 & 58.7 & 82.9 & 84.0 & 74.7 & 95.6 & 95.4 & 96.8 & 99.3 & 96.8 \\ 
        +ATEN (ours)  & $\surd$ & 76.3 & 60.6 & 84.5 & 83.7 & 76.3 & 95.8 & 96.0 & \textbf{100.0} & 99.7 & 97.9 \\ \midrule
        CAiDA & $\surd$ & 70.3 & 55.0 & 83.0 & 80.7 & 72.2 & 95.2 & 95.6 & 98.1 & 99.7 & 97.1 \\ 
        +ATEN (ours) & $\surd$ & 76.1 & 60.3 & 85.1 & 83.5 & 76.3 & \textbf{95.9} & \textbf{96.3} & \textbf{100.0} & 99.7 & \textbf{98.0} \\ \midrule
        DATE & $\surd$ & 75.2 & 60.9 & \textbf{85.2} & 84.0 & 76.3 & 95.6 & 95.7 & 98.1 & \textbf{99.8} & 97.3 \\ 
        +ATEN (ours) & $\surd$ & \textbf{76.7} & 61.6 & \textbf{85.2} & \textbf{84.7} & \textbf{77.1} & \textbf{95.9} & 95.7 & \textbf{100.0} & 99.7 & 97.8 \\ \bottomrule
    \end{tabular}
    \caption{Results on Office-Home and Office-Caltech. The `+ATEN' rows show improvements obtained by plugging ATEN into original methods. SF denotes whether the method follows source-free setting. Best results are  in bold font.}
    \label{tab3}
\end{table*}

\subsection{Main Results}
\textbf{DomainNet}. \tref{tab2} illustrates classification accuracies on the DomainNet dataset. Note that methods end with -best are originally single-source domain adaptation approaches, and we select their single-best results on each target domain for fair comparison. AVG-ENS is a naive ensemble strategy by averaging over outputs from all source models, and is listed as a baseline. 
The results show that our Bi-ATEN achieves superior performance on most of the tasks, except for domains \textbf{pnt} and \textbf{qdr} we are behind PMTrans. This is because PMTrans has access to labeled source data, which helps to overcome the large domain gaps in DomainNet by distribution alignment. Bi-ATEN exhibits clear enhancements than ATEN, especially on the two most challenging tasks \textbf{inf} and \textbf{qdr}. Under such significant domain shift,  bottleneck-classifier pairs learned by Bi-ATEN show better compatibility.
%from the same domain are very likely to mismatch,  leading to biased weight learning process. Bi-ATEN overcomes this issue by effectively assembling the most compatible bottlenecks and classifiers. 
The Train time column compares training time among available source-free methods on target \textbf{clp}. Our methods achieve higher accuracy in considerably less training time.   The Param. column compares trainable parameters of existing open-source methods.  Source-free methods train more parameter as they tune all source models. Larger transformer-based backbones also require heavy computation overheads. Our methods require significantly less trainable parameters to surpass all competing method while following source-free setting, demonstrating the efficacy and agility of our methods.
Another key observation is that all existing MSFDA methods are implemented on ResNet50 backbone due to high computational complexities, largely limiting their performance. Our Bi-ATEN stands out as  the first MSFDA method that introduces large models like SwinTransformer as backbone while maintaining a surprisingly low computation cost. Notably, our Bi-ATEN achieves a remarkable performance improvement of 7\% over the current SOTA in MSFDA task, and additionally demonstrates comparable or even superior performance with source-available domain adaptation methods, strongly supporting the validity and efficiency of the proposed method.

\begin{table}[t]
    \centering
    \small
    \begin{tabular}{lcccc}
        \toprule
        Method & $\rightarrow$clp & $\rightarrow$inf & $\rightarrow$skt & Avg. \\ \midrule
        Bi-ATEN (ours) & \textbf{77.0} & 38.5 & \textbf{64.9} & \textbf{60.1} \\
        \begin{tabular}[l]{@{}l@{}}ATEN (ours) \\ (w/o intra-domain weights)\end{tabular} & 76.6 & 37.2 & 64.6 & 59.5 \\
        w/o alternate training & 75.8 & \textbf{38.6} & 64.1 & 59.5 \\
        w/o $\mathcal{L}_{intra}$ & 76.1 & 35.8 & 63.4 & 58.4 \\			
        w/o $\mathcal{L}_{IM}$ & 75.8 & 38.5 & 63.6 & 59.3 \\ \bottomrule   			
    \end{tabular}
    \caption{Ablation study on three tasks from DomainNet. Best results are in bold font.}
    \label{tab4}
\end{table}

\textbf{Office-Home and Office-Caltech.} \tref{tab3} gives performance improvements obtained by plugging ATEN to existing MSFDA methods. Results show that computing ensemble weights by ATEN brings a maximal 4.1\% overall accuracy boost and hardly any negative effects. 
The combination DATE+ATEN achieves the best accuracy on Office-Home with +0.8\% improvement while more significant boost can be observed on baselines DECISION and CAiDA. On Office-Caltech, CAiDA+ATEN achieves the highest accuracy of 98\%, approaching fully-supervised performance. We notice that accuracies obtained by plugging in ATEN tend to be similar within the same dataset despite the varying baseline performance. This phenomenon indicates that ATEN is able to learn stable ensemble strategies disregarding potential perturbations from origin method, which guarantees fair performance and steady improvements on various baselines. The experiment provides compelling evidence that ATEN is not only effective with fixed backbones but also offers promising enhancements when applied to existing MSFDA methods, suggesting that learning ensemble weights through our ATEN is beneficial.

\begin{figure}[t]
    \centering
    \begin{subfigure}{0.23\textwidth}
        \includegraphics[width=\textwidth]{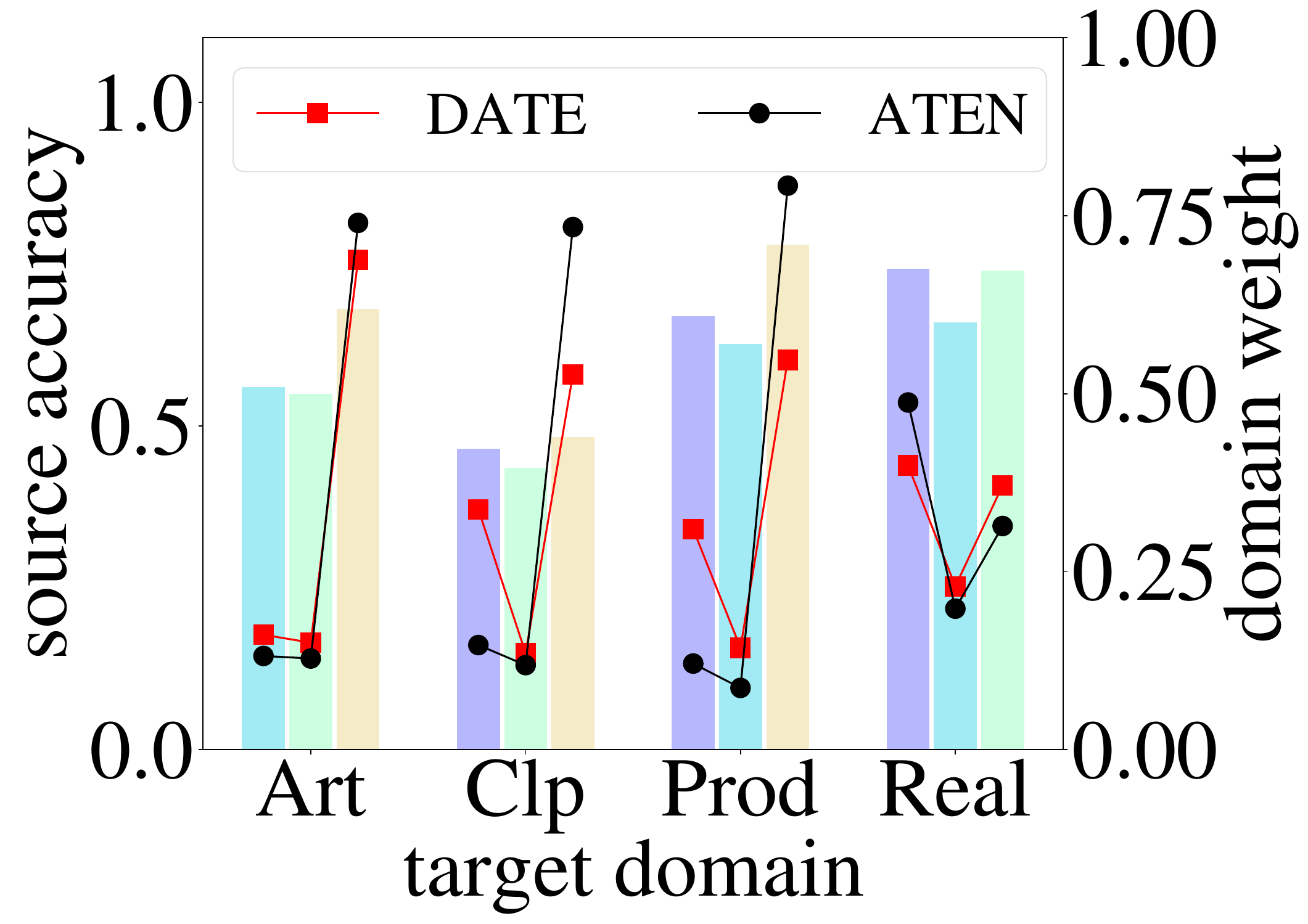}
        \caption{Office-Home}
        \label{fig3_1}
    \end{subfigure}
    \begin{subfigure}{0.23\textwidth}
        \includegraphics[width=\textwidth]{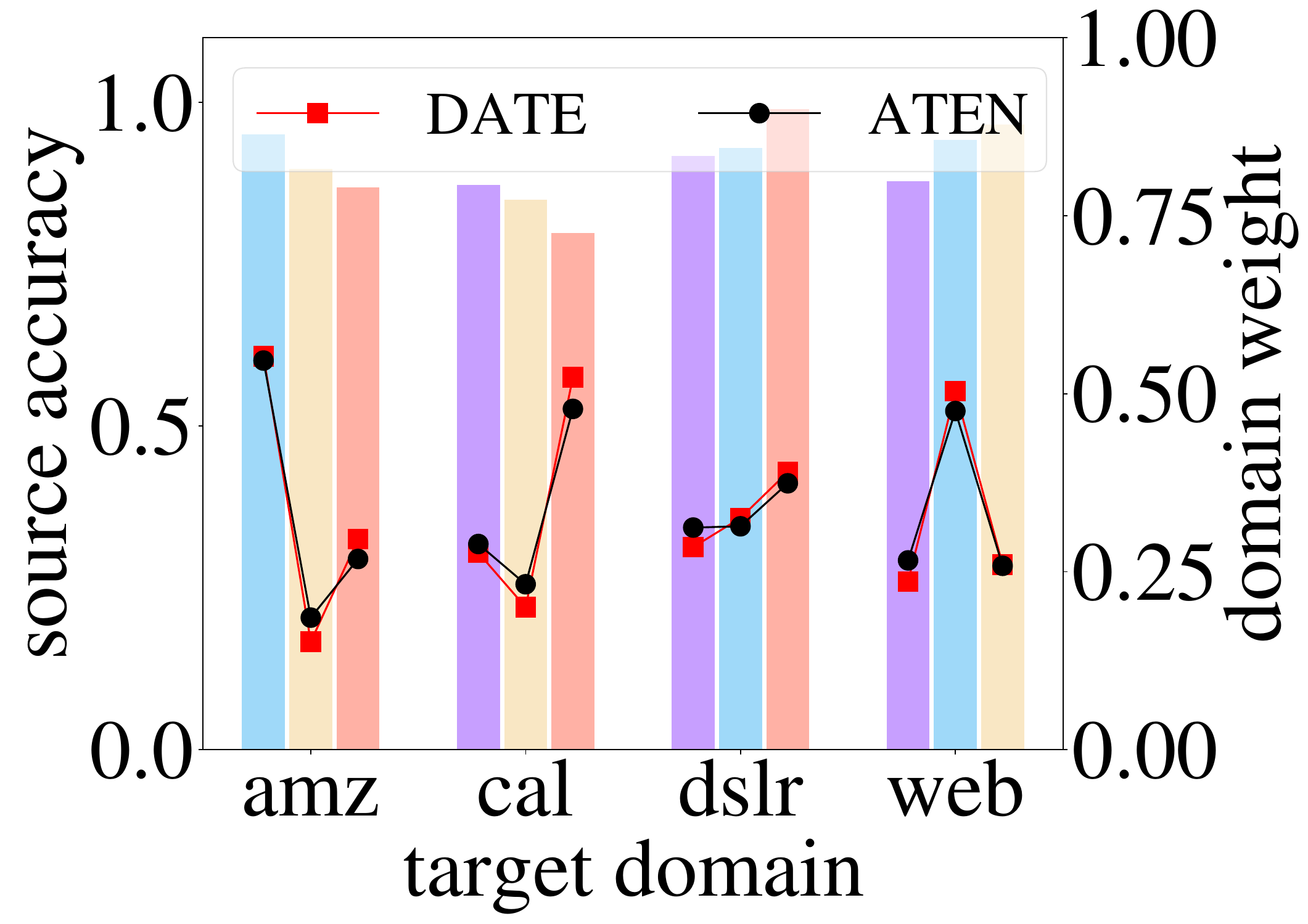}
        \caption{Office-Caltech}
        \label{fig3_2}
    \end{subfigure}
    \caption{Domain-level inter-domain weight comparison. Bars represent source-only accuracies of source models. Lines represent averaged weights assigned to each source.}
    \label{fig3}
\end{figure}
\begin{figure}[t]
    \centering
    \begin{subfigure}{0.23\textwidth}
        \includegraphics[width=\textwidth]{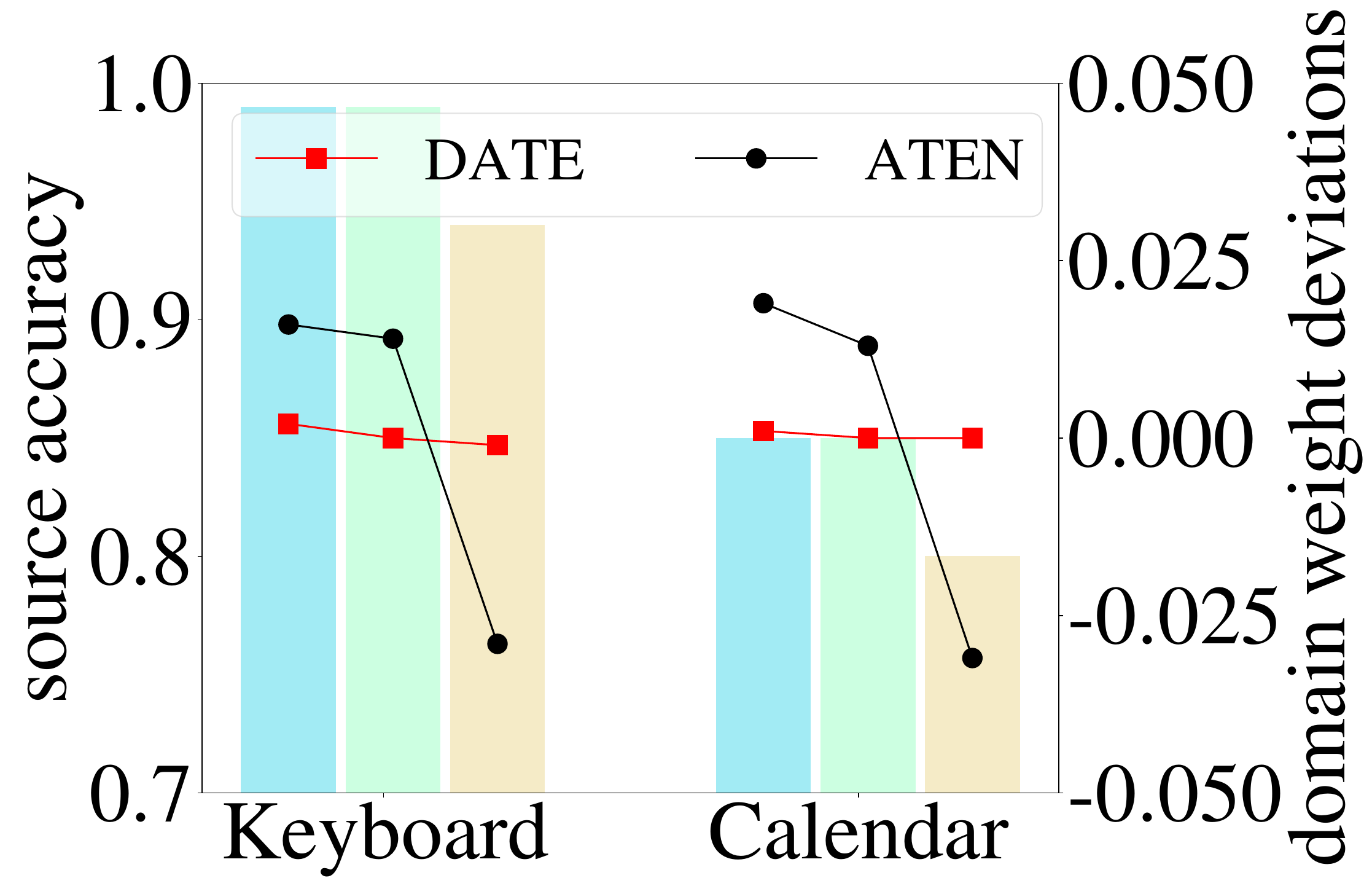}
        \caption{Target: Art.}
        \label{fig4_1}
    \end{subfigure}
    \begin{subfigure}{0.23\textwidth}
        \includegraphics[width=\textwidth]{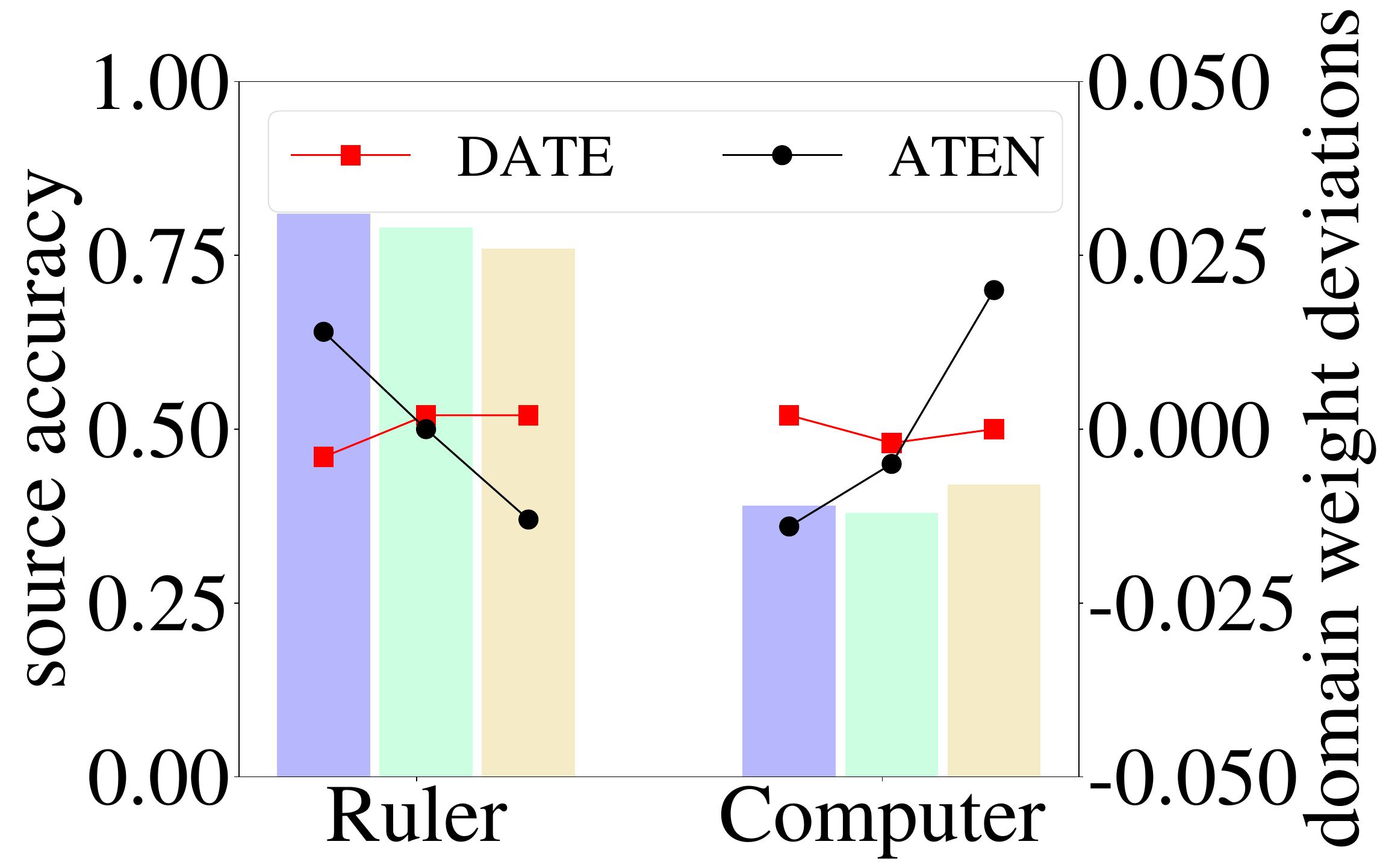}
        \caption{Target: Clp.}
        \label{fig4_2}
    \end{subfigure}
    \caption{Class-level inter-domain weight comparison on Office-Home. Bars represent source accuracy. Lines represent  weight deviations assigned to each source output.}
    \label{fig4}
\end{figure}

\subsection{Analytical Experiments}
\textbf{Ablation study.} \tref{tab4} presents ablation study by removing different modules in our framework, where w/o $\mathcal{L}_{IM}$ is to remove the IM loss in Eq. (15). It can be concluded that all modules contribute positively to our method, and the complete framework Bi-ATEN achieves the best overall accuracy.  The alternate training procedure aims to balance the adaptation performance under both small and large distribution shift by focusing on domain specific bottleneck-classifier pairs in certain epochs. However, this procedure could harm the learning of intra-domain weights under significant domain shift as in task $\rightarrow$\textbf{inf}. Therefore, removing alternate training can lead to slight accuracy increase in these challenging tasks. Removing $\mathcal{L}_{intra}$ brings the largest performance decay, suggesting that learning inappropriate intra-domain weights can harm final outcomes. The IM loss is more effective in easier tasks ($\to$clp, $\to$skt) where well-classified classes might mislead similar classes. On hard tasks ($\to$inf) where most samples are misclassified,
mode collapse rarely occurs thus IM loss is less effective.

\textbf{Weight analysis.} We present a comprehensive analysis on the two types of weights learned in our framework. \fref{fig3} shows that domain-level weights learned by ATEN  aligns well with source model transferabilities and accuracies, and this similarity is comparable to that achieved by DATE. This demonstration emphasizes that ATEN effectively learns domain-consistent inter-domain ensemble weights.

\begin{figure}[t]
    \centering
    \begin{subfigure}{0.23\textwidth}
        \includegraphics[width=\textwidth]{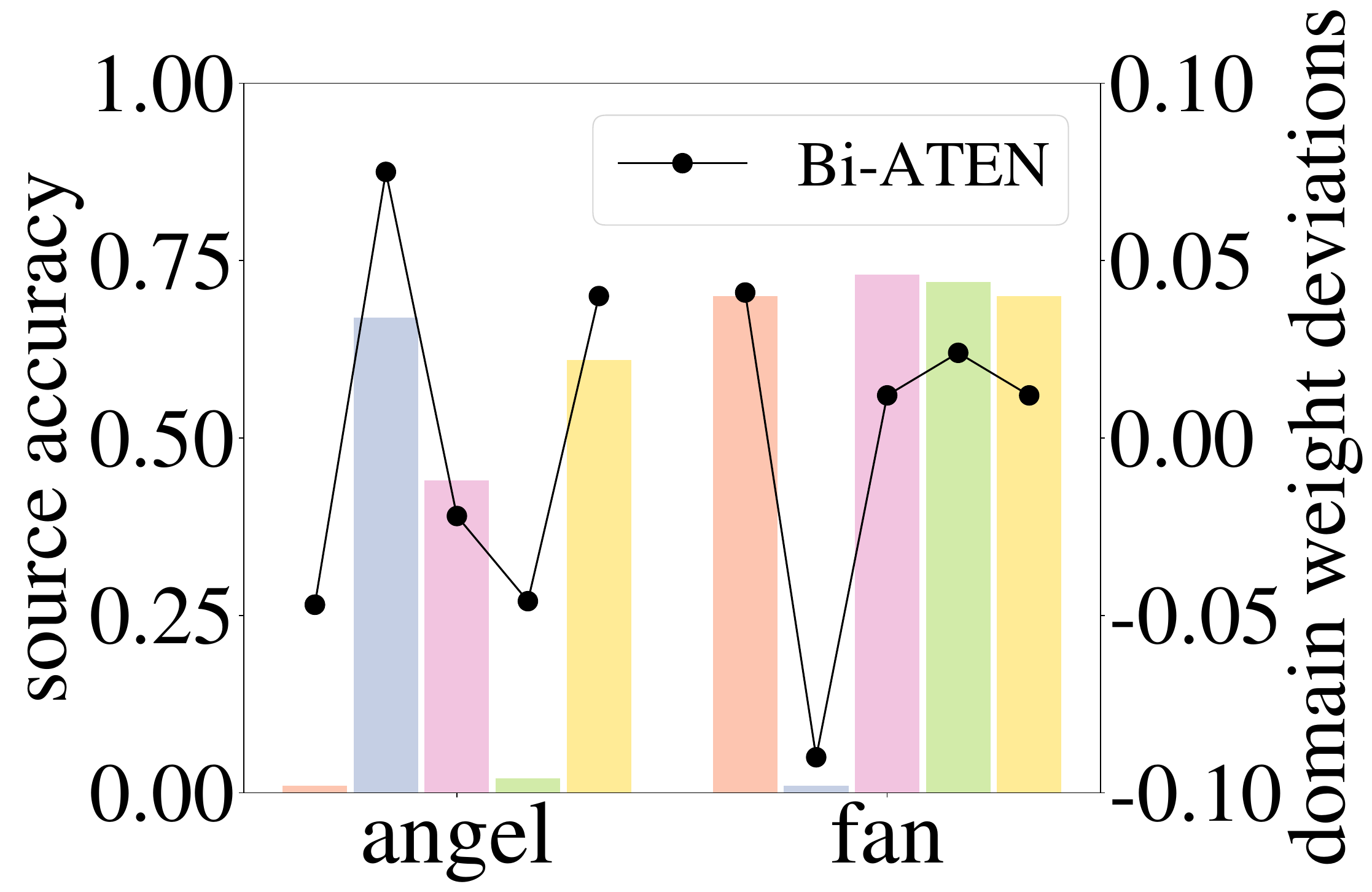}
        \caption{Target: clp.}
        \label{fig5_1}
    \end{subfigure}
    \begin{subfigure}{0.23\textwidth}
        \includegraphics[width=\textwidth]{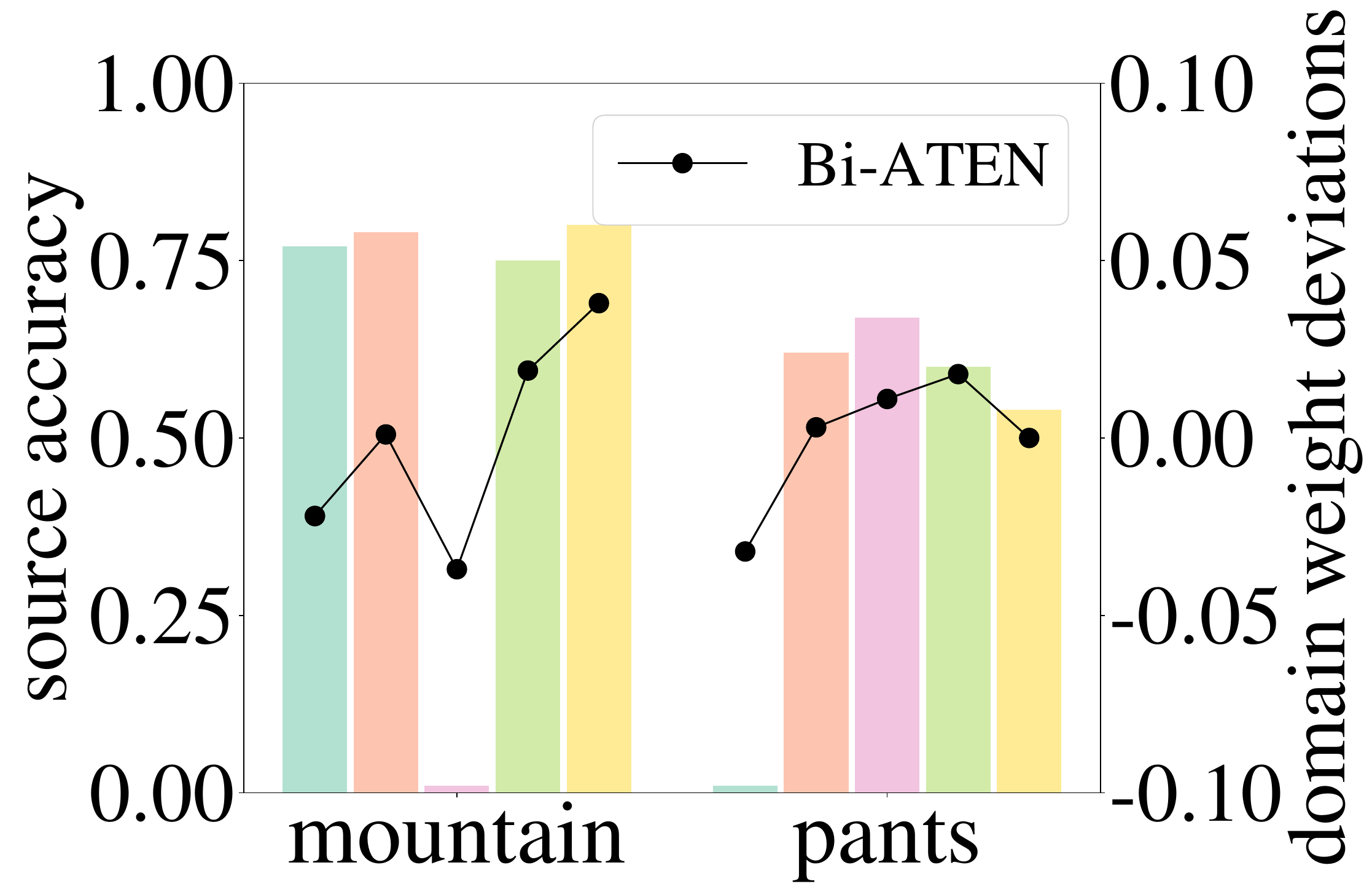}
        \caption{Target: pnt.}
        \label{fig5_2}
    \end{subfigure}
    \begin{subfigure}{0.23\textwidth}
        \includegraphics[width=\textwidth]{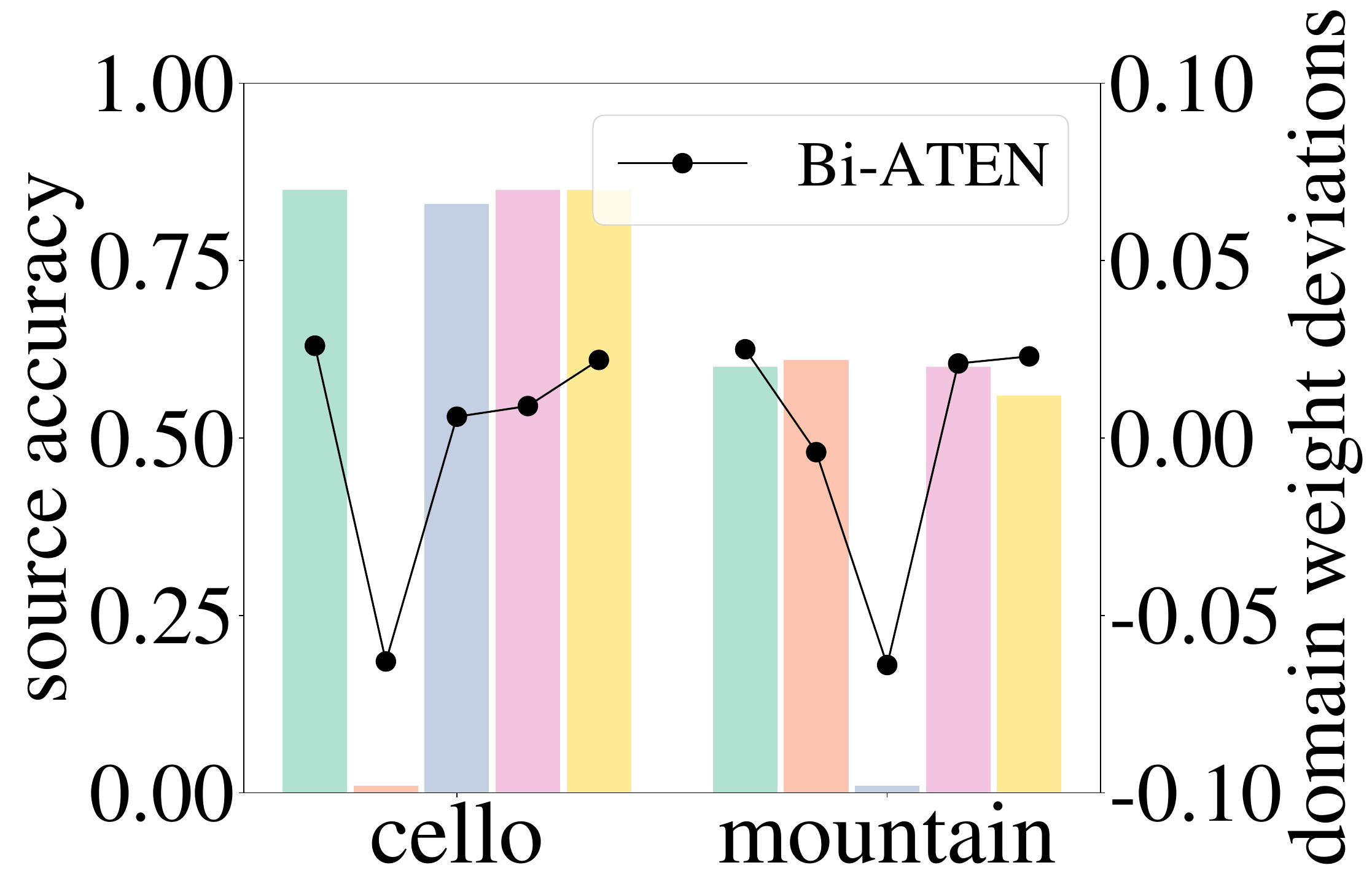}
        \caption{Target: rel.}
        \label{fig5_3}
    \end{subfigure}
    \begin{subfigure}{0.23\textwidth}
        \includegraphics[width=\textwidth]{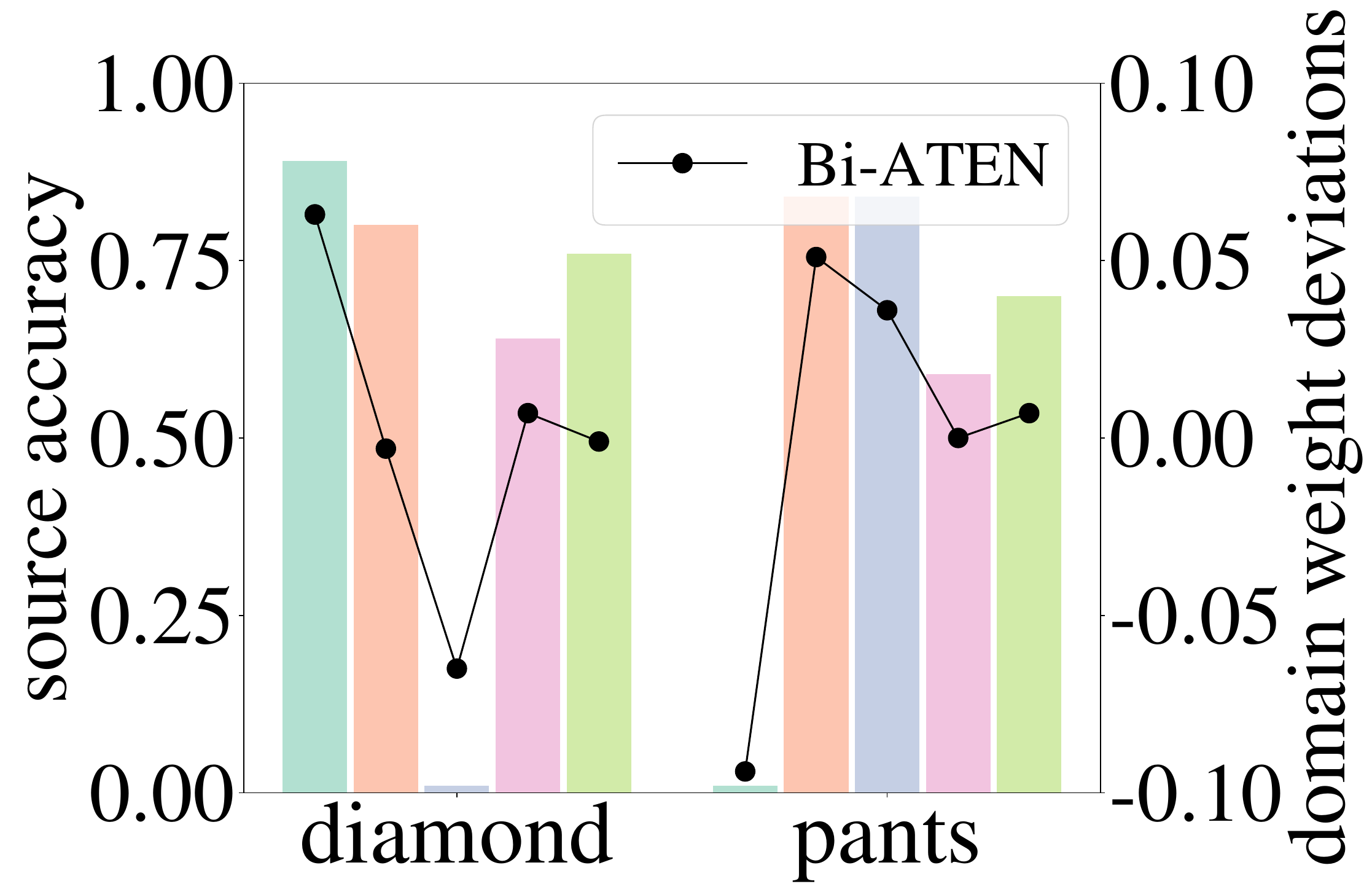}
        \caption{Target: skt.}
        \label{fig5_4}
    \end{subfigure}
    \caption{Class-level inter-domain weights on DomainNet. Bars represent source accuracies and lines represent domain weight deviations assigned to each source output.}
    \label{fig5}
\end{figure}

Limited flexibility of identical inter-domain ensemble weights prevent them from accommodating special instances with unique transfer characteristics, ultimately leading to a decline in performance. 
Our method addresses this by learning tailored inter-domain weights. We examine classes instead of instances for the sake of brevity. \fref{fig4} represents how the class-level inter-domain weights deviates from domain-level weights, showcasing their ability to dynamically adapt to different classes that require distinct transferabilities. In contrast to DATE, which shows limited class-level adaptability, ATEN demonstrates its ability to learn individualized and effective strategies by striving to derive suitable weights customized for each class. However, without intra-class weights, this customization is limited, as the deviations are relatively subtle in \fref{fig4}.
\fref{fig5} provides the results on DomainNet of our full design. Under more significant transferability gap, Bi-ATEN is still able to adapt intelligently to source models with zero transferability by actively reducing their corresponding weights to prevent negative transfer. The tailored weights are deviated more significantly with the help of intra-domain weights. The collaborative evidence presented in \fref{fig3}, \fref{fig4} and \fref{fig5} strongly supports that our method indeed learns weights that are specific to instances and consistent on domains.

\begin{figure}[t]
    \centering
    \begin{subfigure}{0.22\textwidth}
        \includegraphics[width=\textwidth]{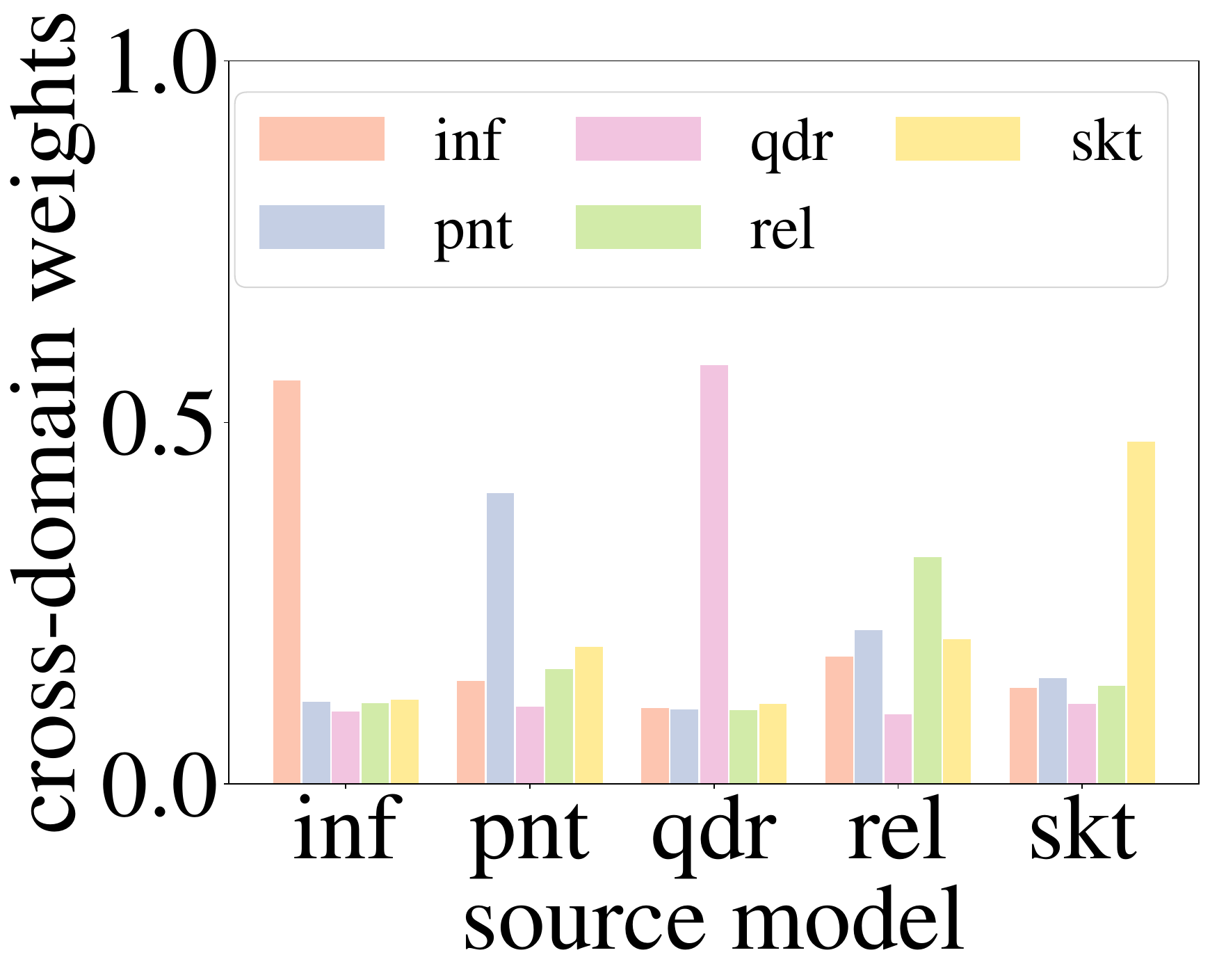}
        \caption{Target: clp.}
        \label{fig6_1}
    \end{subfigure}
    \begin{subfigure}{0.22\textwidth}
        \includegraphics[width=\textwidth]{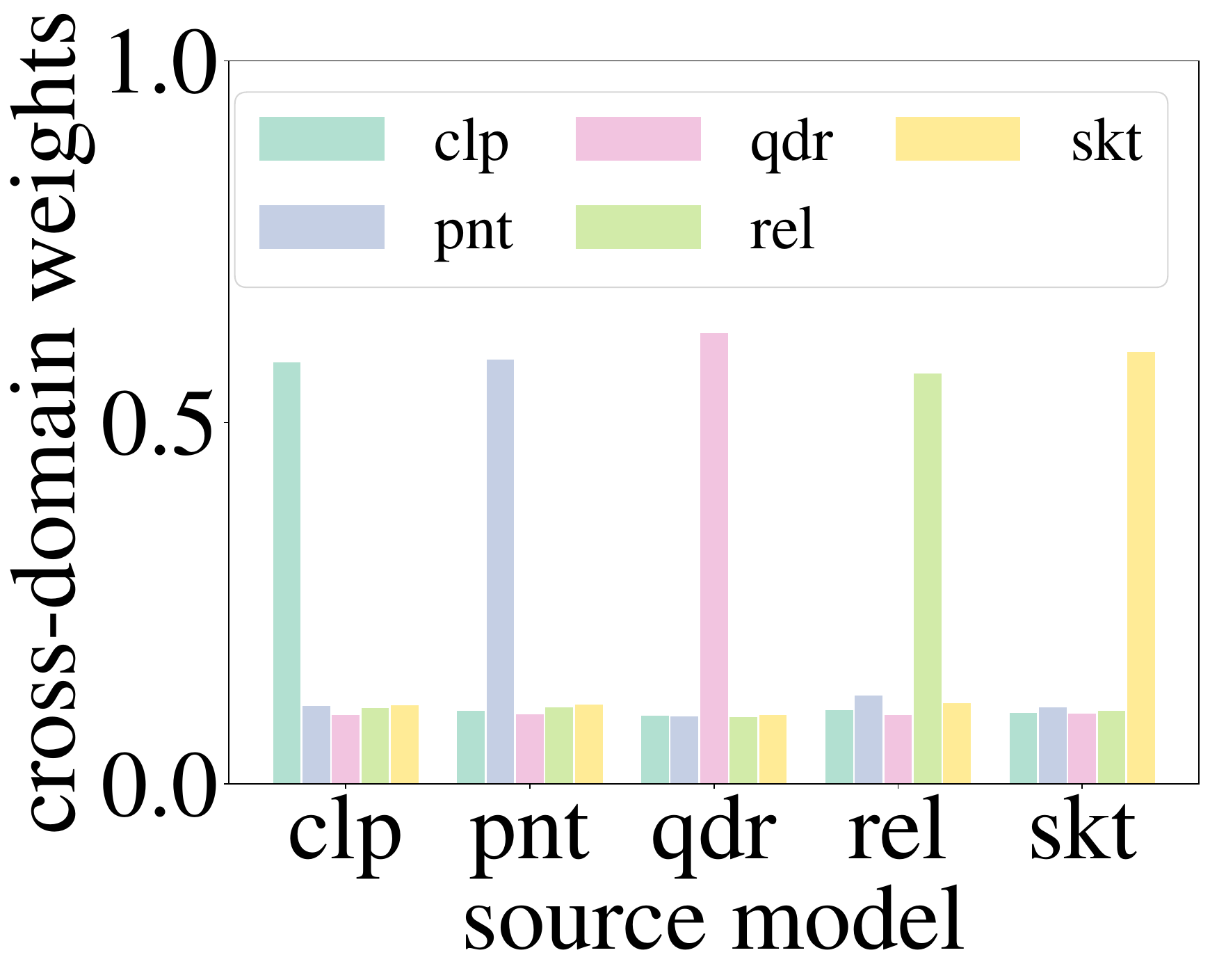}
        \caption{Target: inf.}
        \label{fig6_2}
    \end{subfigure}
    \caption{Intra-domain weights on DomainNet. Bars represent intra-domain weights assigned to each source classifier.}
    \label{fig6}
\end{figure}

\begin{figure}[t]
    \centering
    \begin{subfigure}{0.23\textwidth}
        \includegraphics[width=\textwidth]{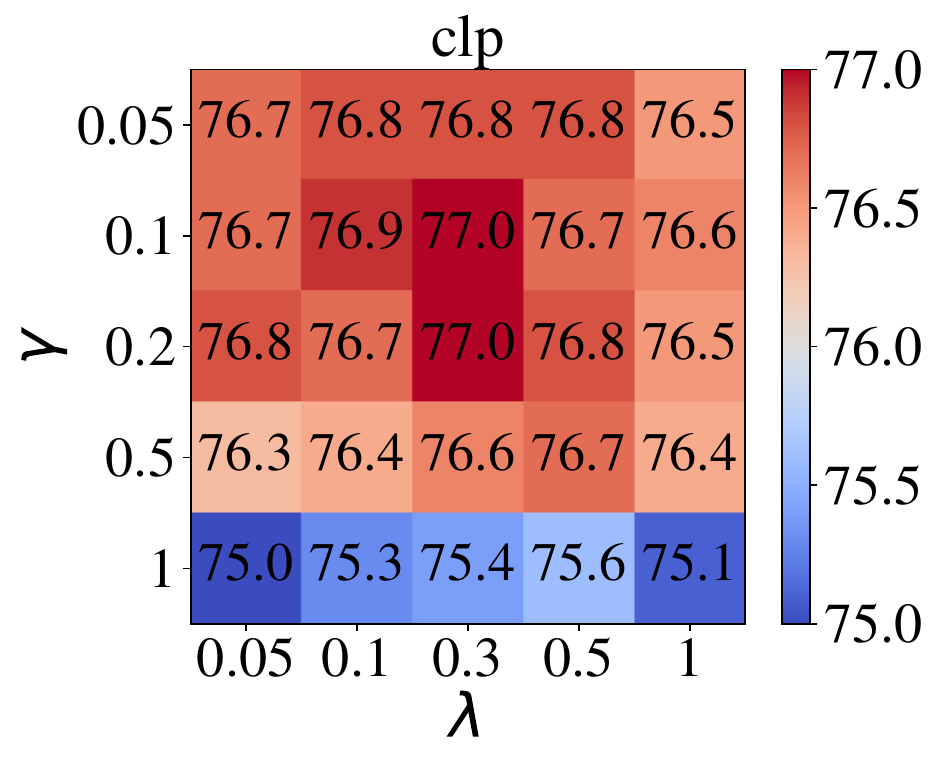}
        \caption{Target: clp.}
        \label{fig7_1}
    \end{subfigure}
    \begin{subfigure}{0.23\textwidth}
        \includegraphics[width=\textwidth]{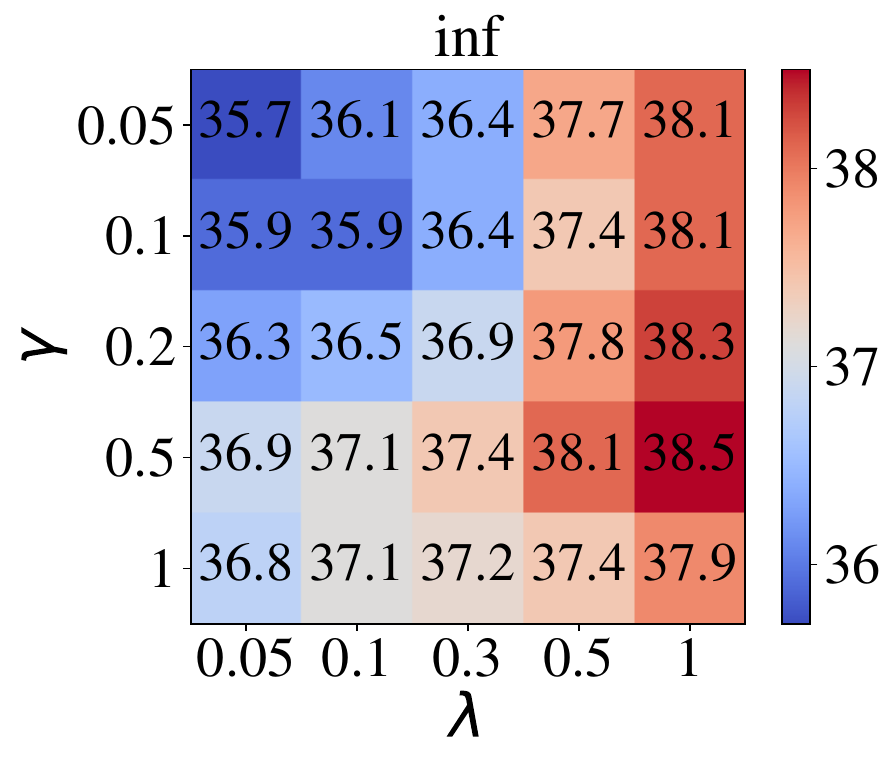}
        \caption{Target: inf.}
        \label{fig7_2}
    \end{subfigure}
    \caption{Hyperparameter analysis on DomainNet. Numbers represent overall accuracy obtained by each hyperparameter combination.}
    \label{fig7}
\end{figure}

Intra-domain weights learned by Bi-ATEN are presented in \fref{fig6}. Each group of weights are corresponding intra-domain weights $\boldsymbol{\alpha^i}$ for the source bottleneck feature. It can be seen that the classifiers from the same domain as bottleneck features receive the majority of attention. However, this attention can also dynamically match more compatible target domains, as exemplified in source \textbf{rel} of \fref{fig6_1}.

\textbf{Hyperparameter analysis.} \fref{fig7} gives accuracies under different hyperparameters in Eq. (15) and Eq. (16). Results show that a large $\gamma$  harms performance, which suggests that overly relying on pseudo labels misguides the weight learning process. For target domains with larger domain gap (target \textbf{inf}), a larger $\lambda$ is needed to constrain the intra-domain weights to avoid negative transfer, as stated in ablation study. Optimal parameter combinations might vary across different target data, but the overall performance is relatively stable.

\section{Conclusion}
This research aims to address the high computation costs associated with existing MSFDA methods. We present a novel framework that prioritizes the learning of instance-specific and domain-consistent ensemble weights, instead of extensively tuning each source model. We achieve this by designing a novel bi-level attention module that effectively learns intra-domain and inter-domain weights. Extensive experiments demonstrate that our methods significantly outperform state-of-the-art methods while requiring considerably lower computation costs.
We believe that our work has the potential to  encourage the exploration of more light-weight approaches to address the challenges posed by MSFDA.

\section{Acknowledgements}
We thank all reviewers for their hard work and thoughtful feedbacks. 
This work was supported in part by the National Natural Science Foundation of China under Grant 62250061, 62176042 and 62173066, and in part by Sichuan Science and Technology Program under Grant 2023NSFSC0483, and in part Sponsored by CAAI-Huawei MindSpore Open Fund.
%\end{document}
\bibliography{aaai24}
%\begin{document}

\newpage
       \twocolumn[
        \centering
        \Large
        \vspace{0.5em}\textbf{Supplementary Material for Agile Multi-Source-Free Domain Adaptation} \\
        \vspace{2.0em}
       ] %< twocolumn

\section{Training Algorithm}
We first apply dynamic-clustering to obtain pseudo labels. Centroid for class $c$ generated by source model $h_s^i$ is obtained by:
\begin{equation}
    \label{eq1}
    \mu_c^i = \frac{\sum_{x_t\in X_t} \delta_c(Softmax(\ddot{y}_t))\phi_t^i}{\sum_{x_t\in X_t} \delta_c(Softmax(\ddot{y}_t))},  
\end{equation}
where $\delta_c(\cdot )$ takes the $c_{th}$ logit. Dynamic centroid for the $m_{th}$ target sample $x_t^m$ of class $c$ is computed by assembling all centroids using instance-specific domain weight $\boldsymbol{\beta^m}$:
\begin{equation}
    \label{eq2}
    \tilde{\mu}_c^m = \sum_{i=1}^n \beta_i^m \mu_c^i.
\end{equation}
Finally, we generate pseudo label for $x_t^m$ by:
\begin{equation}
    \label{eq3}
    y_t = \arg \underset{c}{\max} \; Cosine(\tilde{\phi}_t^m, \ \tilde{\mu}_c^m),
\end{equation}
where $\tilde{\phi}_t^m=\sum_{i=1}^{n} \beta_i^m \phi_t^i $, and that $\phi_t^i$ is bottleneck feature for sample $x_t^m$.

For training of Bi-ATEN, we first obtain intra-domain weight $ \boldsymbol{\alpha^i} $ for bottleneck feature $\phi_t^i$  by:
\begin{align}
\label{eq4}
    & O_t^i = Concat(\{ \theta_{g_s}^j \phi_t^i \}_{j=1}^n, \ \dim=0),\\
    & Sim_t^i = Cosine(\phi_t^i W^{F}, O_t^i W^{O}),\\
    & \boldsymbol{\alpha^i} = Softmax(Sim_t^i).
\end{align}
Assembled output for domain $i$ is then obtained by:
\begin{equation}
    \label{eq5}
    \tilde{y}_t^i=\sum _{j=1}^{n} \alpha_j^i \theta_{g_s}^j \phi_t^i.
\end{equation}
Next we compute inter-domain weights $\boldsymbol{\beta} $ similarly:
\begin{align}
    \label{eq6}
    & \hat{\phi}_t^K = Concat(\{\phi_t^i W^F\}_{i=1}^n, \ \dim=0),\\
    & \hat{\phi}_t^Q = Concat(\{\phi_t^i\}_{i=1}^n, \ \dim=1)W^{QF},\\
    & \boldsymbol{\beta} = Softmax(Cosine(\hat{\phi}_t^Q, \hat{\phi}_t^K)).
\end{align} 
Finally, we obtain target classification result by:
\begin{equation}
    \label{eq7}
    \ddot{y}_t=\sum _{i=1}^{n} \beta_i \tilde{y}_t^i.
\end{equation}

\begin{algorithm}[t]
    \caption{Alternate Training of Bi-ATEN}
    \label{alg1}
    \textbf{Input}: Unlabeled target data $X_t$, alternate interval $d_{alter}$, iterations to complete an epoch $epoch\_iter$, maximum iteration number $max\_iter$, source-trained models $\{h_s^i\}_{i=1}^{n}$.\\
    \textbf{Output}: Instance-customized $\boldsymbol{\alpha}$ and $\boldsymbol{\beta}$, trained source bottlenecks $k_s$.
    \begin{algorithmic}[1] %[1] enables line numbers
    \STATE Let $iter=0$, $epoch=0$.
    \WHILE{$iter < max\_iter$}
    \STATE Sample a batch of target inputs $x_t$ from $X_t$.
    \IF {$\ iter\mod epoch\_iter ==  0$}
    \STATE Update pseudo labels $Y_t$ by Eq. (\ref{eq1}), Eq. (\ref{eq2}) and Eq. (\ref{eq3}).
    \STATE Let $epoch=epoch+1$.
    \ENDIF
    \IF {$\ epoch \mod d_{alter} \neq  0$}
    \STATE Obtain intra-domain weights $\boldsymbol{\alpha}$ by Eq. (4), Eq. (5) and Eq. (6).
    \ELSE
    \STATE Set $\boldsymbol{\alpha}$ as one-hot weights where only the domain-indexed position is 1.\\
    \ENDIF 
    \STATE Obtain classification result $\ddot{y_t}$ by Eq. (7), Eq. (8), Eq. (9), Eq. (10) and Eq. (11).
    \STATE Retrieve pseudo labels $y_t$ for $x_t$ from $Y_t$.
    \STATE Update model parameters by Eq. (15).
    \STATE Let $iter=iter+1$.
    \ENDWHILE
    \STATE \textbf{return} solution
    \end{algorithmic}
\end{algorithm}

Training losses are defined by:
\begin{equation}
    \label{eq11}
    \mathcal{L}_{intra} = \sum_{i=1}^{n} \mathcal{L}_{IM}(Softmax(\tilde{y}_t^i)),
\end{equation}
\begin{equation}
    \label{eq12}
    \mathcal{L}_{inter} = \gamma CE(\ddot{y}_t, y_t) + \mathcal{L}_{IM}(Softmax(\ddot{y}_t)),
\end{equation}
\begin{equation}
    \label{eq13}
    \mathcal{L} = \mathcal{L}_{inter} + \lambda \mathcal{L}_{intra},
\end{equation}
where $\gamma$ and $\lambda$ are hyperparameters. Overall optimization objective is given by:
\begin{equation}
    \label{eq14}
    \boldsymbol{\alpha }, \ \boldsymbol{\beta }, \ \theta_{k_s} = \arg\min \;\mathcal{L}.
\end{equation}
where $\theta_{k_s}$ is parameter of source bottlenecks.

\begin{table}[t]
    \centering
    \footnotesize
    \caption{Optimal hyperparameters on DomainNet.}
    \label{tab1}
    \begin{tabular}{cccccccc}
    \toprule
     & $\rightarrow$clp & $\rightarrow$inf & $\rightarrow$pnt & $\rightarrow$qdr & $\rightarrow$rel & $\rightarrow$skt \\ \midrule
    $\lambda$ & 0.2 & 1 & 1 & 1 & 1 & 0.2 \\
    $\gamma$ & 0.1 & 0.5 & 0.1 & 0.1 & 0.1 & 0.1 \\ 
    epoch & 30 & 30 & 30 & 500 & 30 & 30 \\ 
    initial lr & 0.02 & 0.02 & 0.02 & 0.05 & 0.02 & 0.02 \\ \bottomrule
    \end{tabular}
\end{table}

\begin{figure*}[t]
    \centering
    \begin{subfigure}{0.24\textwidth}
        \includegraphics[width=\textwidth]{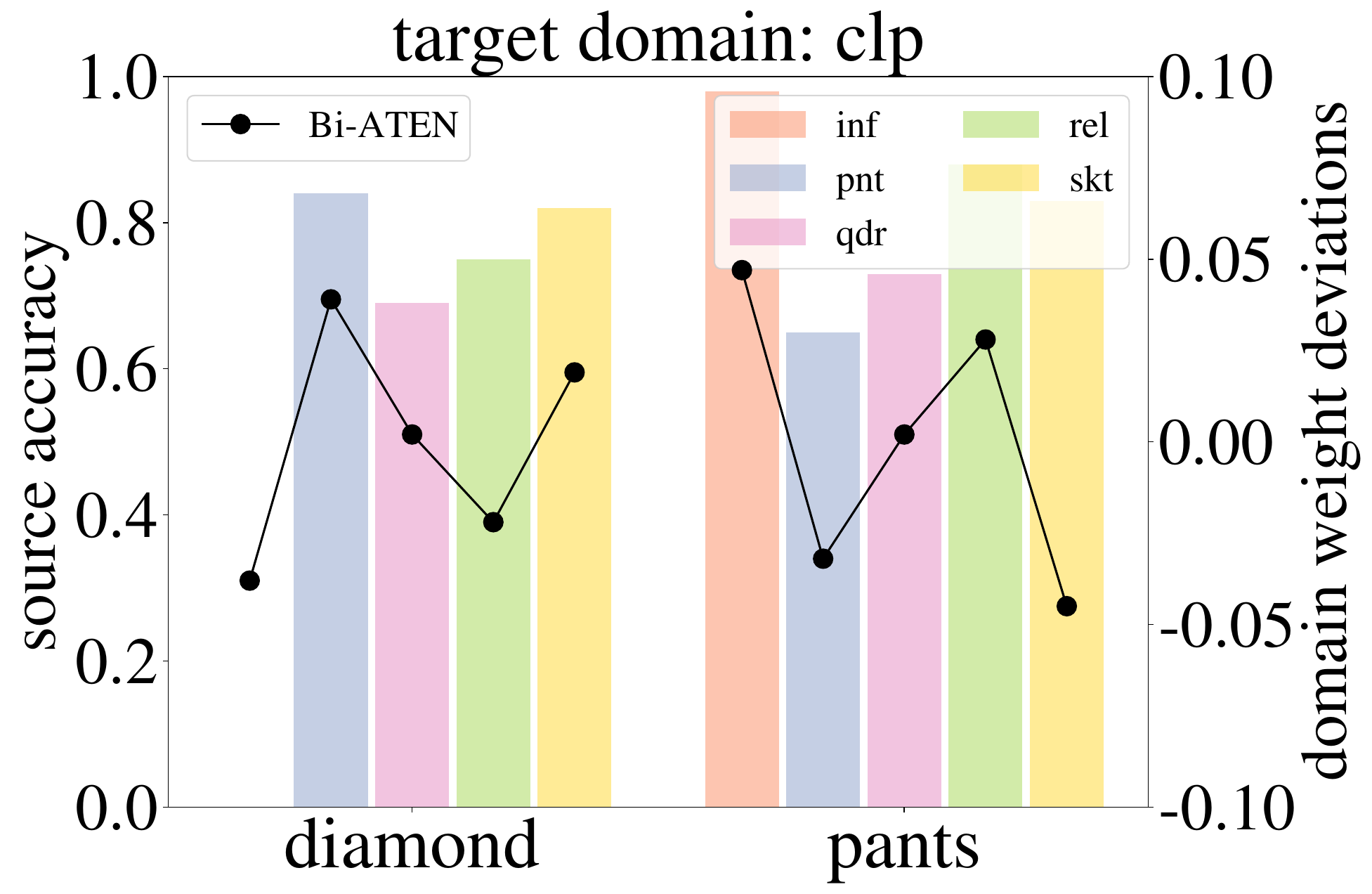}
        \caption{Target: clp.}
        \label{fig1_1}
    \end{subfigure}
    \begin{subfigure}{0.24\textwidth}
        \includegraphics[width=\textwidth]{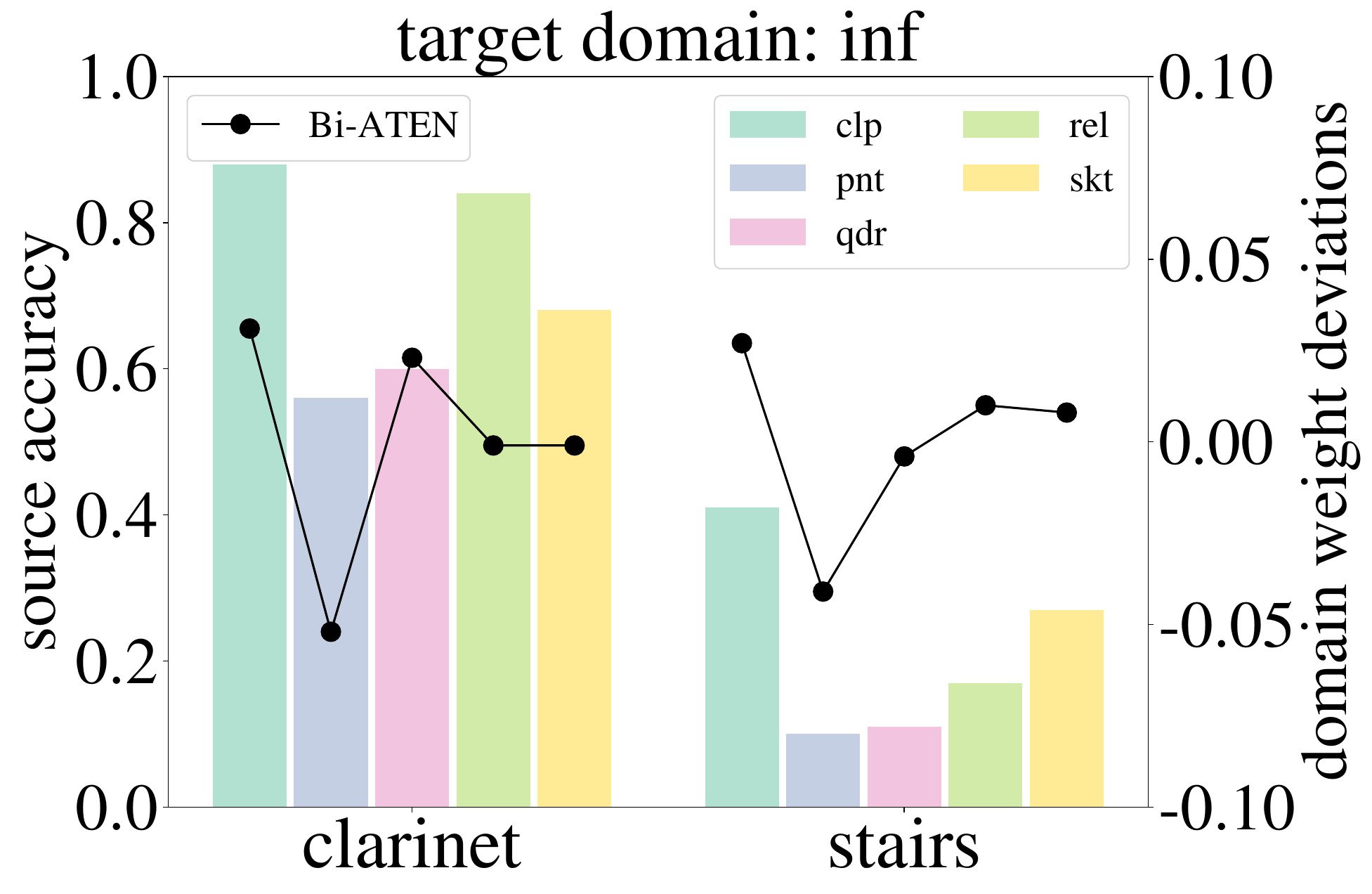}
        \caption{Target: inf.}
        \label{fig1_2}
    \end{subfigure}
    \begin{subfigure}{0.24\textwidth}
        \includegraphics[width=\textwidth]{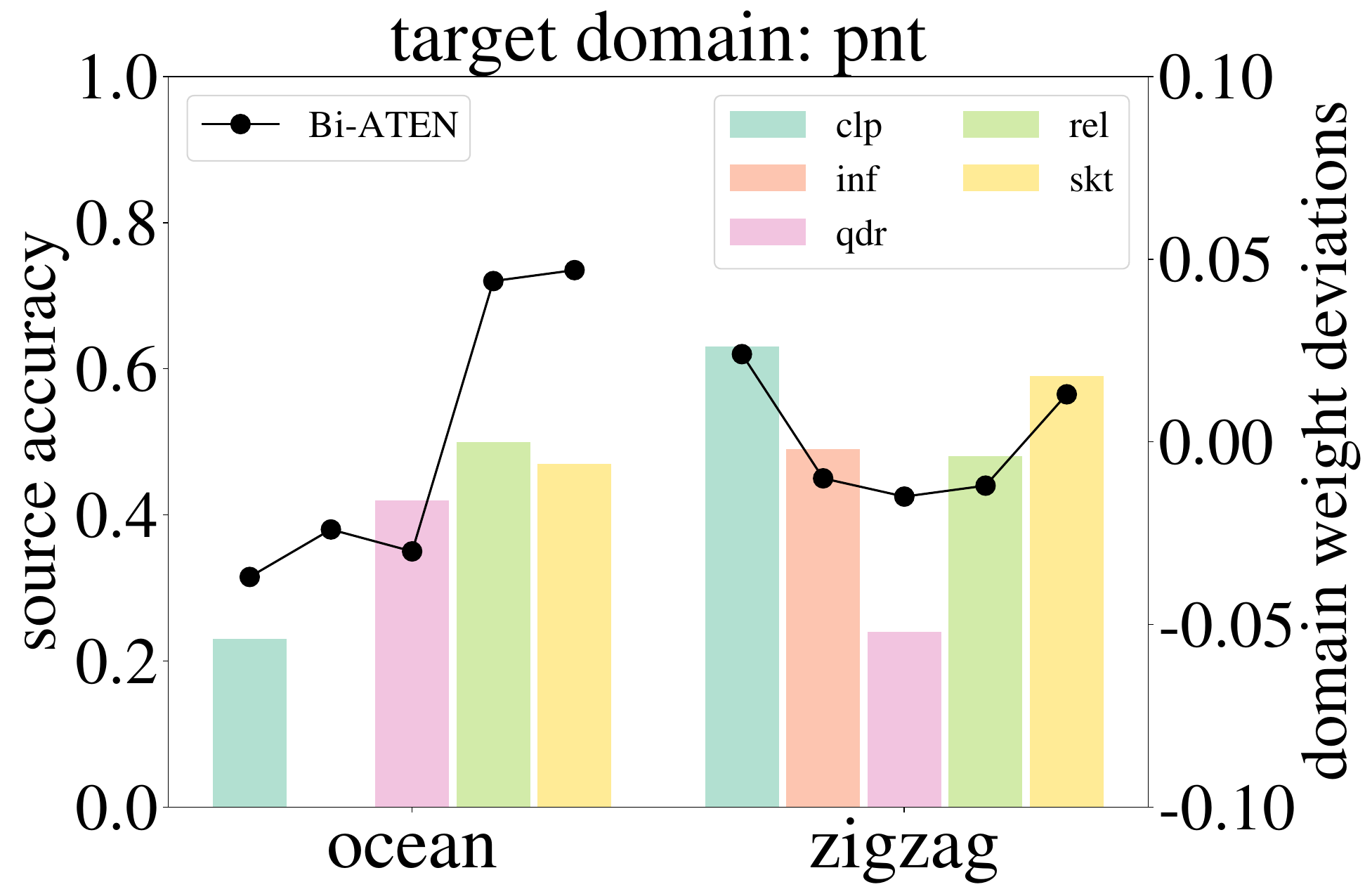}
        \caption{Target: pnt.}
        \label{fig1_3}
    \end{subfigure}
    \begin{subfigure}{0.24\textwidth}
        \includegraphics[width=\textwidth]{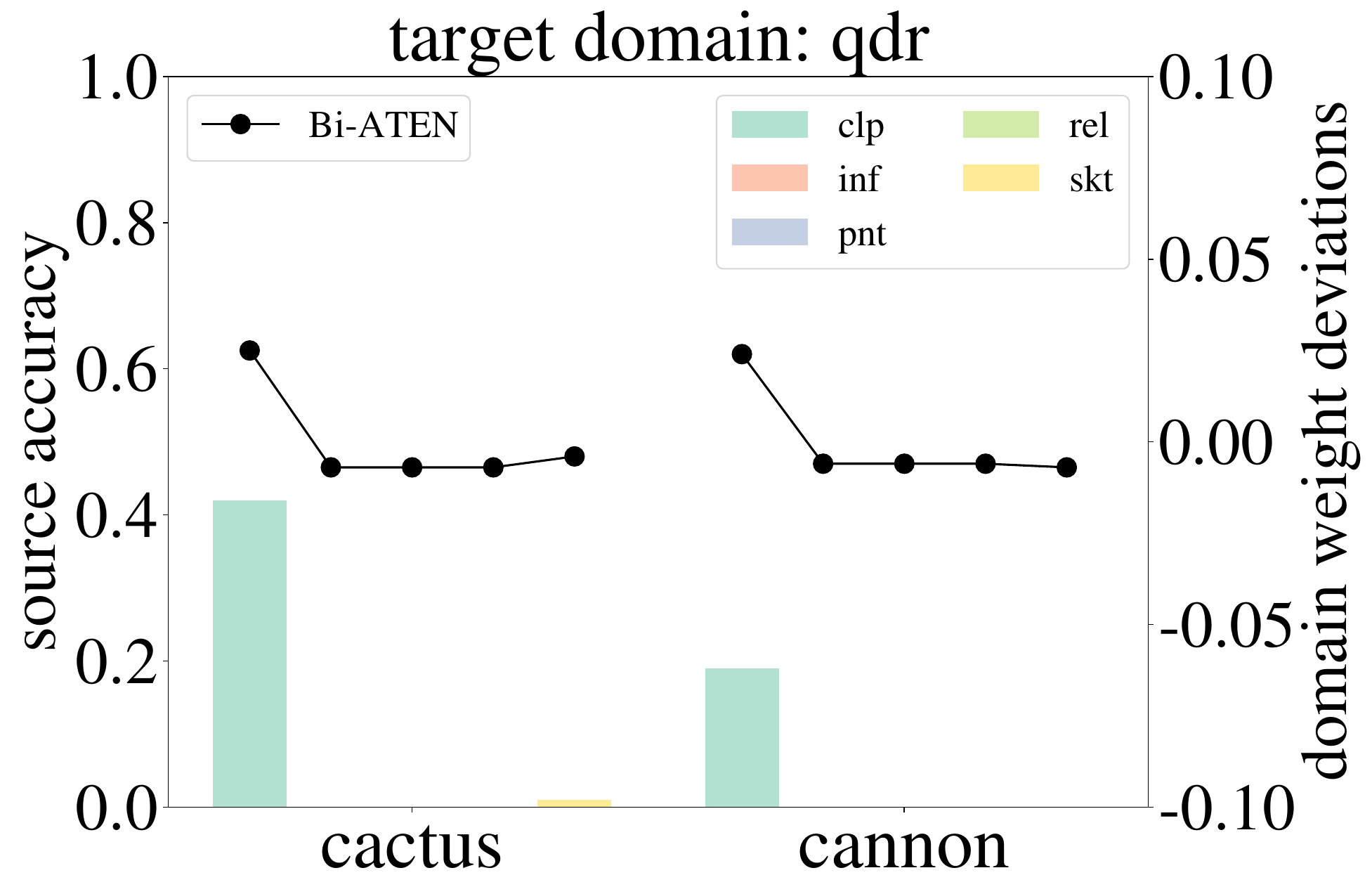} %%
        \caption{Target: qdr.}
        \label{fig1_4}
    \end{subfigure}
    \begin{subfigure}{0.24\textwidth}
        \includegraphics[width=\textwidth]{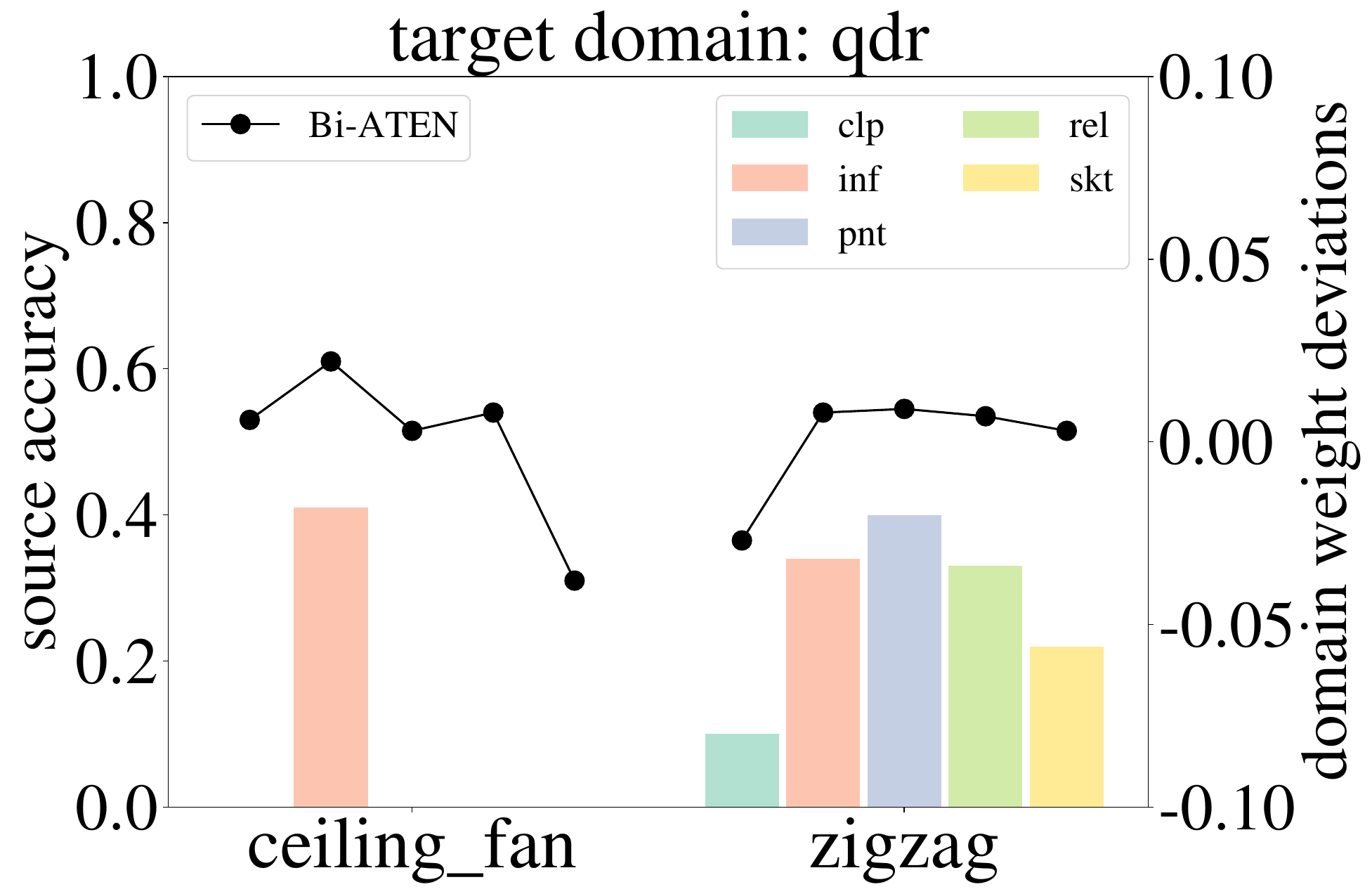} %%
        \caption{Target: qdr.}
        \label{fig1_5}
    \end{subfigure}
    \begin{subfigure}{0.24\textwidth}
        \includegraphics[width=\textwidth]{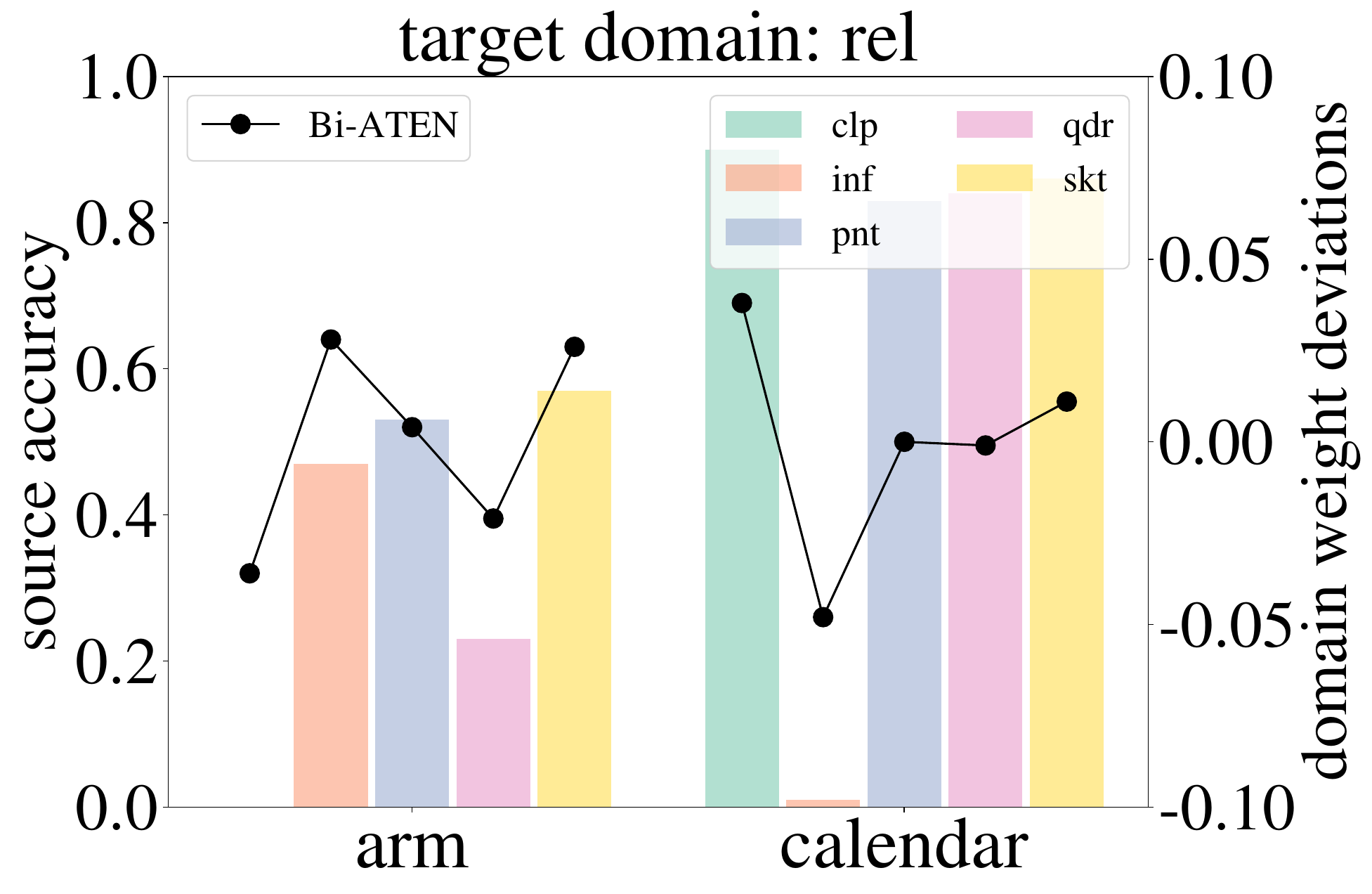}
        \caption{Target: rel.}
        \label{fig1_6}
    \end{subfigure}
    \begin{subfigure}{0.24\textwidth}
        \includegraphics[width=\textwidth]{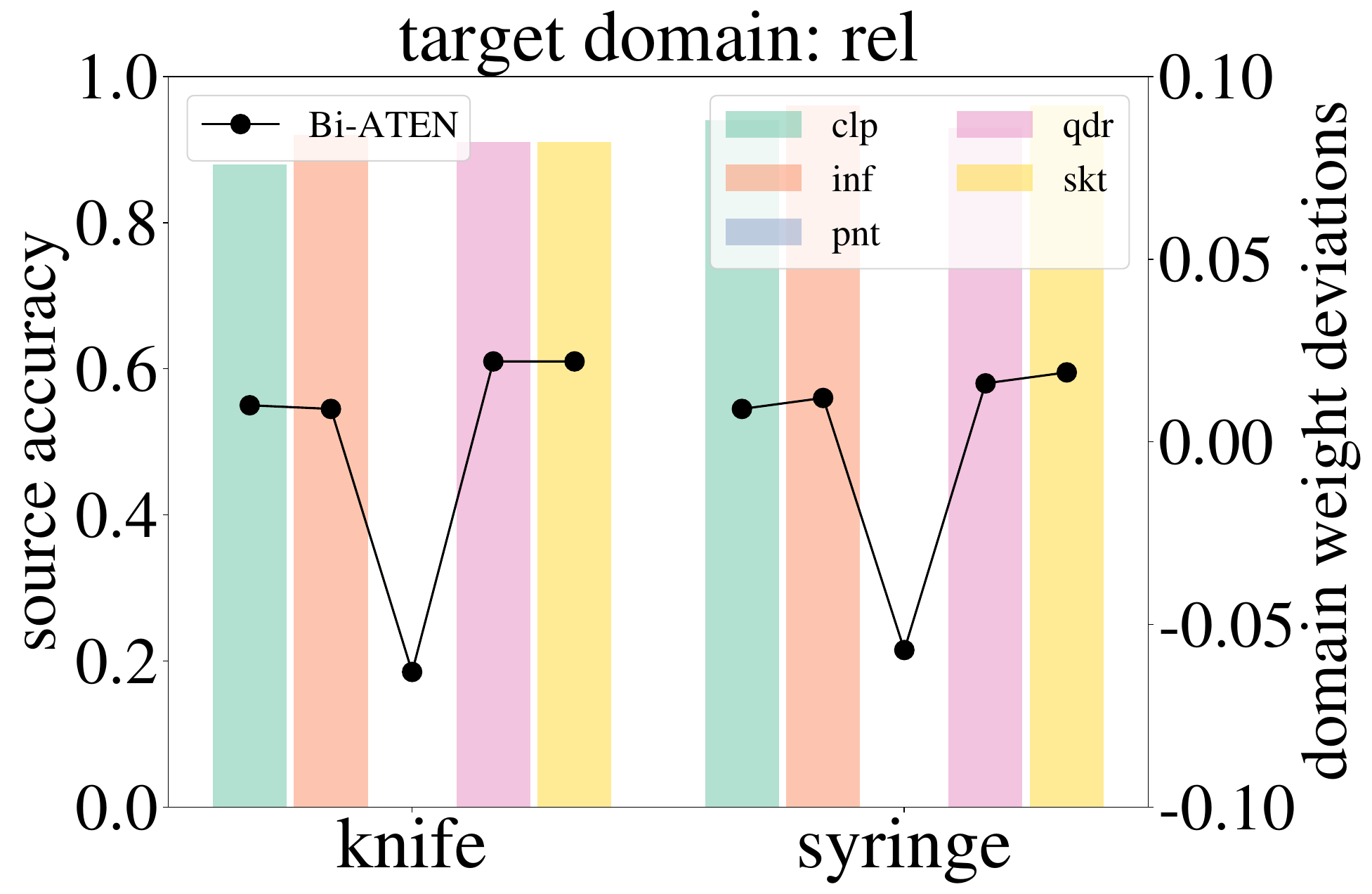}
        \caption{Target: rel.}
        \label{fig1_7}
    \end{subfigure}
    \begin{subfigure}{0.24\textwidth}
        \includegraphics[width=\textwidth]{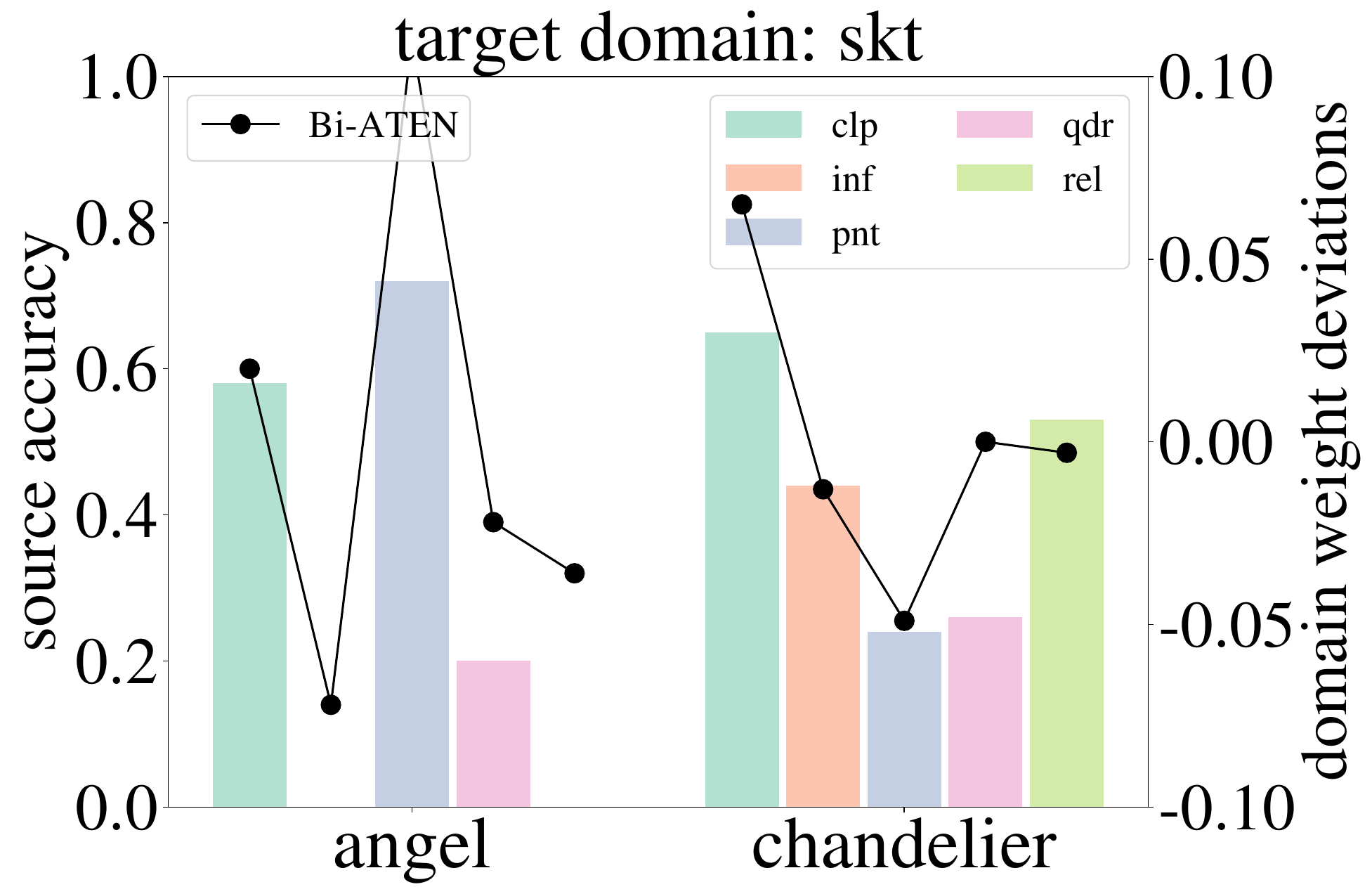}
        \caption{Target: skt.}
        \label{fig1_8}
    \end{subfigure}
    \caption{Class-level inter-domain weights on DomainNet. Bars represent source accuracies on the class. Lines represent inter-domain weight deviations assigned to each source output.}
    \vspace{-5pt}
    \label{fig1}
\end{figure*}

The proposed alternate training paradigm is described in Algorithm \ref{alg1}. For every $d_{alter}$ epochs, intra-domain weights are directly set to one-hot vectors that only considers classifier from the same domain as bottleneck feature.

\section{Multihead Bi-ATEN}
We further improve Bi-ATEN by applying multi-head attention. In order to cope with output vector with arbitrary dimension (arbitrary), each head takes the entire input vector instead of evenly split the vector for different heads. The final similarity between query $Q$ and key $K$ is the average of all heads, which can be denoted as:
\begin{equation}
    \label{eq1_}
    Cosine(Q, K)=Average(Head_i(Q,K)), 
\end{equation}
where $Head_i$ is similarity score computed by Eq. (5) and Eq. (10), but with different transform matrices $W_i^O$, $W_i^F$ and $W_i^{QF}$ in different heads. This design allows computing more detailed attentions via different embeddings, enhancing the perception ability of the module.

\begin{figure}[t]
    \centering
    \begin{subfigure}{0.23\textwidth}
        \includegraphics[width=\textwidth]{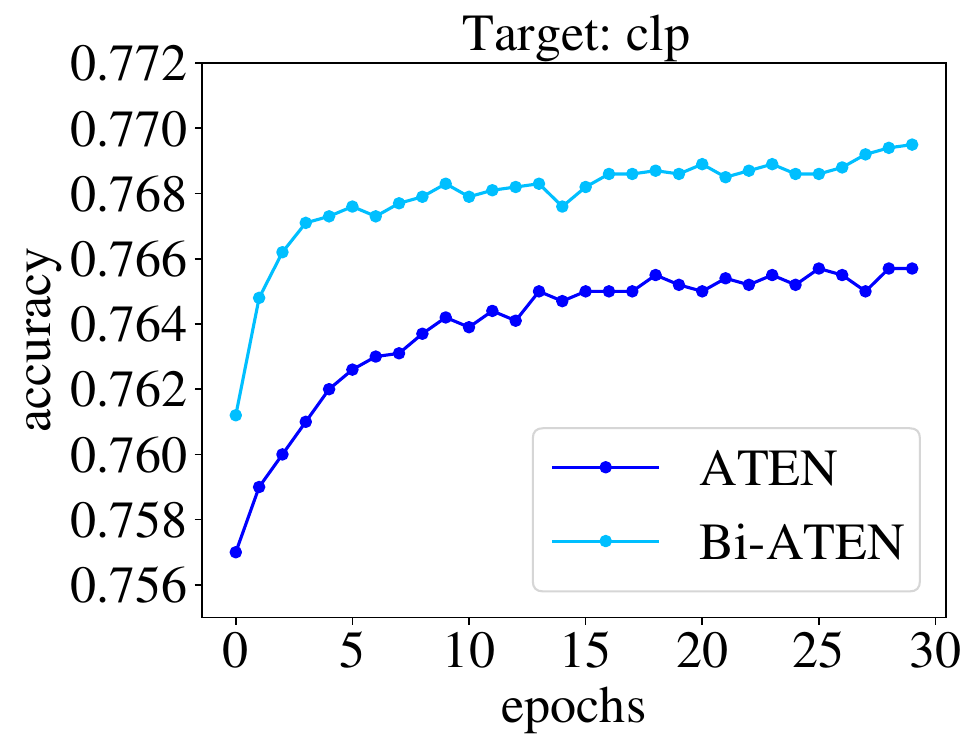}
        \caption{Target: clp.}
        \label{fig2_1}
    \end{subfigure}
    \begin{subfigure}{0.23\textwidth}
        \includegraphics[width=\textwidth]{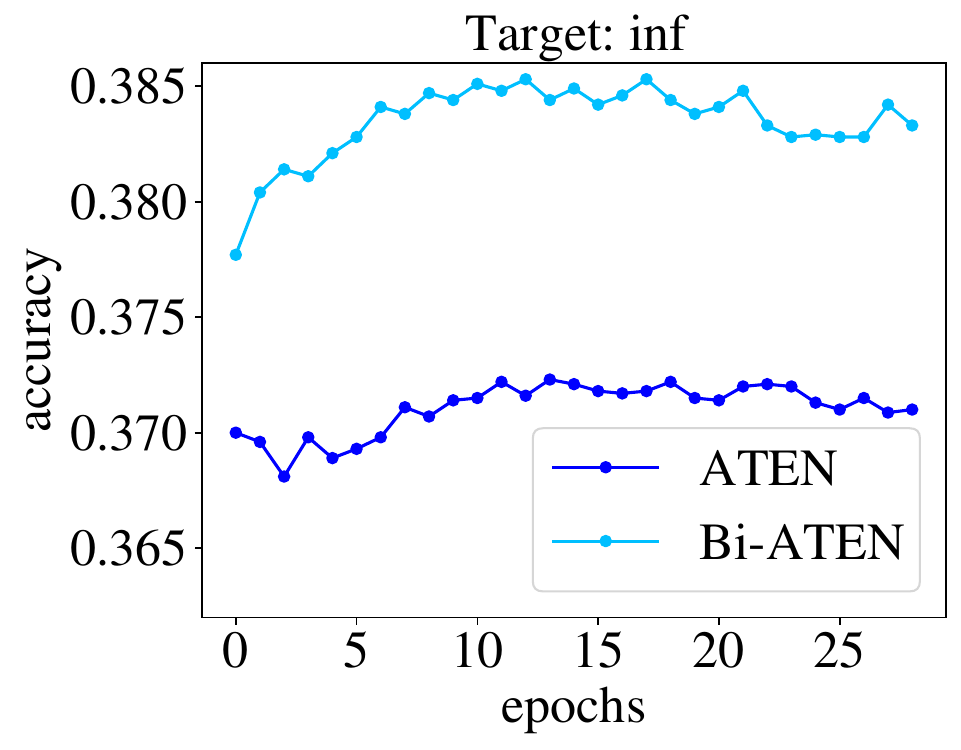}
        \caption{Target: inf.}
        \label{fig2_2}
    \end{subfigure}
    \caption{Convergence analysis on DomainNet.}
    \vspace{-10pt}
    \label{fig2}
\end{figure}

\section{Implementation Details}
We implement all experiments under Python==3.8 on an NVIDIA GeForce RTX 3090 GPU. 
\subsection{Pre-training Source Models}
Each source model includes a bottleneck layer consisting of a single FC layer $k_s \in \mathbb{R}^{d_{backbone}\times d_{k}}$ followed by a batch normalization layer~\cite{ioffe2015batch}, where ${d_{backbone}}$ is raw feature dimension from backbone, and $d_k$ is set to 256 for all experiments. For DomainNet, AdamW~\cite{loshchilov2017decoupled} optimizer with initial learning rate 5e-5 is adopted along with a cosine decay learning rate scheduler as in \cite{liu2021swin}. For Office-Home and Office-Caltech, SGD optimizer with initial learning rate 0.01 is adopted. We use cross entropy loss with label smoothing as training loss on all datasets.
\subsection{ATEN and Bi-ATEN}
Dimension of query and key embeddings ($d_{emb}$) is set to 512 for Bi-ATEN and 2048 for ATEN. Number of heads is set to 4 for both modules. $d_{alter}$ in Algorithm \ref{alg1} is set to 2 for Bi-ATEN. For all datasets, we adopt SGD optimizer with  a cosine decay learning rate scheduler as in \cite{liu2021swin}. Batch size for Office-Home and Office-Caltech is set to 32 and that for DomainNet is 64. Since no training of source backbones is required for Bi-ATEN, we pre-compute backbone features for all instances so that no forward propagation on source backbones is needed during training and inferencing, thus greatly boosting training and inference speed. Hyperparameters, epochs to train and initial learning rate that produce our results on DomainNet are given by \tref{tab1}.

\section{Class-Level Inter-Domain Weights}
We hereby provide more examples on learned inter-domain weights on DomainNet. \fref{fig1} gives example on each of the six domains and various classes. The learned weights strongly support that our Bi-ATEN can learn customized instance-specific ensemble weights over various scenarios, proving the generalizability and efficacy of Bi-ATEN.

\section{Convergence Analysis}
\fref{fig2} shows convergence analysis of ATEN and Bi-ATEN on DomainNet. It can be concluded that Bi-ATEN consistently outperforms ATEN over all examined epochs. On the challenging target domain \textbf{inf}, Bi-ATEN exceeds ATEN by a large margin, demonstrating the necessity and efficacy of  intra-domain weights that introduce more compatible bottleneck-classifier combinations than solely adopting domain-specific pairs.

%\bibliography{aaai24}

\end{document}